%% file: main.tex
\documentclass[10pt,table,twocolumn,letterpaper]{article}

\usepackage{cvpr}              

\input{preamble}

\definecolor{cvprblue}{rgb}{0.21,0.49,0.74}
\usepackage[pagebackref,breaklinks,colorlinks,allcolors=cvprblue]{hyperref}

\def\paperID{13061} 
\def\confName{CVPR}
\def\confYear{2025}

\title{ClearSight: Visual Signal Enhancement for Object Hallucination Mitigation in Multimodal Large Language Models}

\author{Hao Yin \quad Gunagzong Si \quad Zilei Wang\thanks{Corresponding Author} \\
University of Science and Technology of China\\
{\tt\small \{yinhnavi, guangzongsi\}@mail.ustc.edu.cn, zlwang@ustc.edu.cn}
}

\begin{document}
\maketitle
\input{sec/0_abstract} 
\input{sec/1_introduction}
\input{sec/2_related_work}
\input{sec/3_motivation}

\input{sec/4_visual_neglect}
\input{sec/5_method}

\input{sec/6_experiment}
\input{sec/7_conclusion}

\section*{Acknowledgements}
This work is supported by the National Natural Science Foundation of China under Grant 62176246. This work is
also supported by Anhui Province Key Research and Development Plan (202304a05020045),  Anhui Province Natural Science Foundation (2208085UD17) and National Natural Science Foundation of China under Grant 62406098.

{
    \small
    \bibliographystyle{ieeenat_fullname}
    \bibliography{main}
}

\input{sec/X_suppl}

\end{document}

%% file: preamble.tex
\usepackage{tcolorbox}
\usepackage{hanging}
\usepackage{booktabs}
\usepackage{multirow}
\usepackage{xcolor}
\usepackage{stfloats}
\usepackage{pifont}

%% file: sec/0_abstract.tex
\begin{abstract}
Contrastive decoding strategies are widely used to mitigate object hallucinations in multimodal large language models (MLLMs). By reducing over-reliance on language priors, these strategies ensure that generated content remains closely grounded in visual inputs, producing contextually accurate outputs. Since contrastive decoding requires no additional training or external tools, it offers both computational efficiency and versatility, making it highly attractive. However, these methods present two main limitations: (1) bluntly suppressing language priors can compromise coherence and accuracy of generated content, and (2) processing contrastive inputs adds computational load, significantly slowing inference speed. To address these challenges, we propose Visual Amplification Fusion (VAF), a plug-and-play technique that enhances attention to visual signals within the model’s middle layers, where modality fusion predominantly occurs. This approach enables more effective capture of visual features, reducing the model’s bias toward language modality. Experimental results demonstrate that VAF significantly reduces hallucinations across various MLLMs without affecting inference speed, while maintaining coherence and accuracy in generated outputs. The code is available at \url{https://github.com/ustc-hyin/ClearSight}.
\end{abstract}

%% file: sec/1_introduction.tex
\section{Introduction}

In recent years, MLLMs
\cite{blip2,llava,coop,minigpt,llava1.5,shikra} 
have achieved remarkable progress in the intersecting fields of computer vision and natural language processing, and have been widely applied in tasks such as image captioning and visual question answering. However, these models often encounter the issue of "object hallucination"
\cite{li2023evaluating,liu2023mitigating,gunjal2023detecting,lovenia2023negative}
in practical applications, where the generated textual descriptions do not match the actual objects in the image. This problem highlights an over-reliance on unimodal priors (especially language priors)
\cite{yan2023overcoming,zhibo2023overcoming,han2022visual,wu2022overcoming}
during inference, posing potential risks in high-precision applications such as medical diagnosis
\cite{wang2023chatcad,hu2023advancing}
and autonomous driving~\cite{mai2023llm,liu2023llm,chen2023driving,wu2023embodied}.

To address object hallucination
\cite{agrawal2016analyzing,goyal2017making,agarwal2020towards,biten2022let}
, several Contrastive Decoding strategies have been introduced in recent years. Among these, the Visual Contrastive Decoding (VCD) method has shown promise in reducing hallucinations by contrasting output distributions from both original and perturbed visual inputs, thus mitigating the model's excessive reliance on language priors~\cite{gupta2022swapmix,niu2021counterfactual}. Notably, contrastive decoding methods do not require additional training or external tools, offering both computational efficiency and versatility, which has garnered them significant attention. However, these methods present two main limitations:

\begin{tcolorbox}[title=\textit{Limitations of Contrastive Decoding}, colback=blue!10, colframe=blue!65!black, boxrule=0.5mm]
\begin{itemize}
    \item \textit{While reducing over-reliance on language priors, these methods may compromise the coherence and accuracy of generated content.}
    \item \textit{Contrastive decoding necessitates separate processing of the original and contrastive inputs, which considerably increases inference time.}
\end{itemize}
\end{tcolorbox}

To address these shortcomings, we hope to propose a training-free method that can effectively reduces hallucinations without compromising content quality or inference speed. Our saliency analysis of the model’s attention maps reveals that biases toward language in generated content do not arise from an overemphasis on language signals but rather from insufficient attention on visual information during modality fusion. Based on this insight, we introduce a novel, plug-and-play technique to mitigate hallucinations: Visual Amplification Fusion (VAF).

Our analysis indicates that modality fusion in MLLMs primarily occurs within the middle layers. VAF specifically amplifies visual signals at these middle layers, enabling the model to capture more distinctive visual features during fusion, which in turn reduces false descriptions in generated text. This technique not only strengthens the model's visual representations but also retains the beneficial influence of language priors, thus preserving content quality. Furthermore, by eliminating the need to process contrastive samples, VAF maintains inference speed. 

Experimental results validate the effectiveness of the VAF method. Across multiple object hallucination benchmarks, VAF demonstrated notable performance gains, with improvements of approximately 3\% on POPE and 7\% on MME. In terms of coherence and accuracy of generated responses, VCD caused a roughly 19\% decrease on NoCaps, while VAF maintained content quality without negative impacts. Additionally, VCD reduced inference speed by 50\%, whereas VAF had virtually no effect on inference speed.

In summary, the main contributions are as follows:
\begin{itemize}
    \item We identify the negative impacts of contrastive decoding methods on both the quality of generated content and model inference speed.
    \item We analyze the modality fusion mechanism in MLLMs, highlighting its insufficient attention to visual information.
    \item We introduce the VAF method, which effectively mitigates the object hallucination problem while maintaining inference speed, coherence, and accuracy.
    \item We demonstrate the significant performance improvements of the VAF method across multiple object hallucination benchmarks.
\end{itemize}

%% file: sec/2_related_work.tex
\section{Related work}

\subsection{Multimodal Large Language Models}

The development of MLLMs
\cite{mplug,funny,llavamed,gpt4roi}
has advanced from BERT-based decoders to LLM-based architectures
\cite{llama,llama2,llama3,qwen,vicuna,stanford}
, enabling improved multimodal relationship capture
\cite{fuyu,llavanext,internvl,mimic}
. Models like BLIP-2~\cite{blip2} and miniGPT-4~\cite{minigpt} incorporate a Q-Former mechanism, which enhances the alignment between visual and textual inputs, allowing for more precise interactions across modalities. InstructBLIP~\cite{instructblip} builds on this approach by adding task-specific instructions, which improve the model’s understanding of context-sensitive visual semantics. LLaVA~\cite{llava} and Qwen-VL~\cite{qwenvl} utilize simpler linear projection techniques that streamline the alignment process, resulting in improved overall performance on vision-language tasks. However, hallucination issues persist across MLLMs, posing a significant challenge that requires further research.

\subsection{Contrastive Decoding Strategies}

In recent years, Contrastive Decoding
\cite{opera,hacl,huo2024selfintrospectivedecodingalleviatinghallucinations,contrastive}
has emerged as a technique to improve generative model accuracy through contrastive judgment, widely employed to address hallucinations in generated content. For instance, Visual Contrastive Decoding (VCD)~\cite{vcd} contrasts output distributions derived from original and distorted visual inputs, effectively reducing the over-reliance on statistical bias and unimodal priors, two essential causes of object hallucinations. Similarly, Instruction Contrastive Decoding (ICD)~\cite{icd} works by comparing distributions derived from standard and disrupted instructions, thereby removing hallucinated concepts from the original distribution. These contrastive methods help ground generated content closely to visual inputs, resulting in contextually accurate outputs. However, despite these advancements, contrastive decoding faces two primary limitations: slower inference speed and reduced coherence in generated content. To overcome these limitations, we propose the VAF method, which achieves effective hallucination reduction while preserving both inference speed and content coherence.

%% file: sec/3_motivation.tex
\section{Preliminary and Motivation}

In \cref{subsec: contrastive decoding}, we illustrate the working mechanism of contrastive decoding to mitigate hallucinations, using Visual Contrastive Decoding as an example. In \cref{subsec: limitations}, we analysis two main drawbacks of this approach: its potential to disrupt the coherence and accuracy of generated content, and its tendency to slow down model inference.

\subsection{Contrastive Decoding}
\label{subsec: contrastive decoding}
We consider a MLLM parametrized by $\theta$. The model takes as input a textual query $x$ and a visual input $v$, where $v$ provides contextual visual information to assist the model in generating a relevant response $y$ to the textual query. The response $y$ is sampled auto-regressively from the probability distribution conditioned on the query $x$ and the visual context $v$. Mathematically, this can be formulated as:
\begin{equation}
    \begin{aligned}
        y_t & \sim p_\theta\left(y_t \mid v, x, y_{<t}\right) \\
            & \propto \exp \operatorname{logit}_\theta\left(y_t \mid v, x, y_{<t}\right)
    \end{aligned}
\end{equation}
where $y_t$ denotes the token at time step $t$, and $y_{<t}$ represents the sequence of generated tokens up to the time step $(t-1)$.

\begin{figure}[h]
  \centering
   \includegraphics[width=0.98\linewidth]{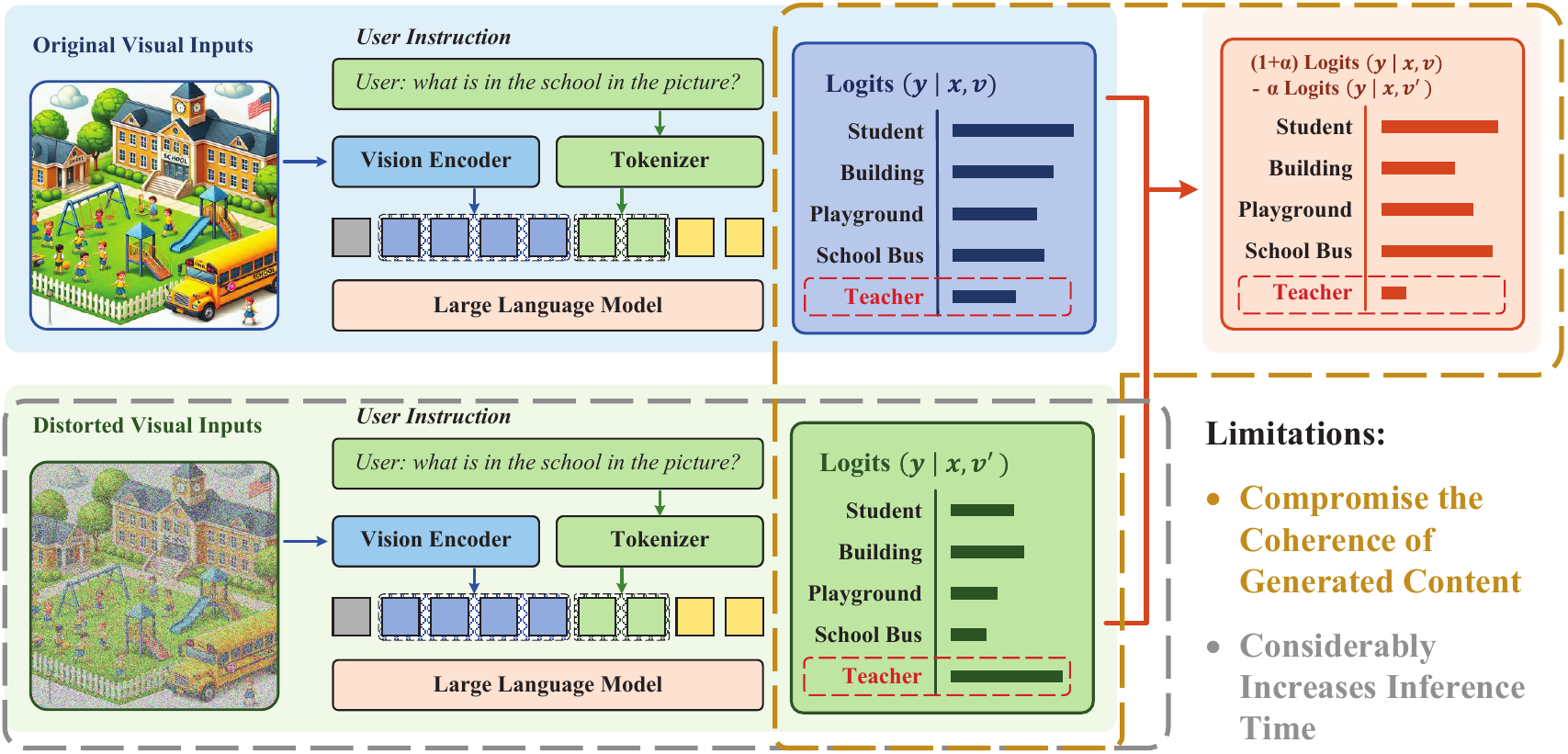}
   \caption{\textbf{Illustration of Visual Contrastive Decoding.} The hallucinated object "Teacher" is suppressed by contrasting with an output distribution prone to hallucinations. This method has two main drawbacks: (1) additional processing of distorted visual inputs greatly increases inference time; (2) subtracting the language prior disrupts content coherence.}
   \label{fig: vcd limitation}
\end{figure}

To mitigate the issue of object hallucination in MLLMs, contrastive decoding techniques can be applied. Here, we present Visual Contrastive Decoding (VCD) as a representative approach, shown in \cref{fig: vcd limitation}. Specifically, given a textual query $x$ and a visual input $v$, the model generates two distinct output distributions: one conditioned on the original $v$ and the other on the distorted visual input $v^{\prime}$, which is derived by applying pre-defined distortions (i.e., Gaussian noise mask) to $v$. Then, a new contrastive probability distribution is computed by exploiting the differences between the two initially obtained distributions. The new contrastive distribution $p_{v c d}$ is formulated as:
\begin{equation}
    \begin{aligned}
        & p_{v c d}\left(y \mid v, v^{\prime}, x\right) = \operatorname{softmax} \Big[ \operatorname{logit}_\theta\left(y \mid v, x\right) + \\
        & \qquad \alpha \cdot \Big(\operatorname{logit}_\theta\left(y \mid v, x\right) - \operatorname{logit}_\theta\left(y \mid v^{\prime}, x\right)\Big) \Big],
    \end{aligned}
\end{equation}
where larger $\alpha$ values indicate a stronger amplification of differences between the two distributions ( $\alpha=0$ reduces to regular decoding). Essentially, VCD serves as a corrective mechanism, reducing hallucinations by contrasting against a distribution predisposed to favoring them.

\subsection{Limitations of Contrastive Decoding}
\label{subsec: limitations}

As contrastive decoding methods do not require training or external tools, they offer high computational efficiency and generalizability, attracting significant attention in academia. However, these methods still have two major drawbacks: a reduction in the quality of generated content and slower inference speed.

\begin{figure}[h]
  \centering
   \includegraphics[width=0.85\linewidth]{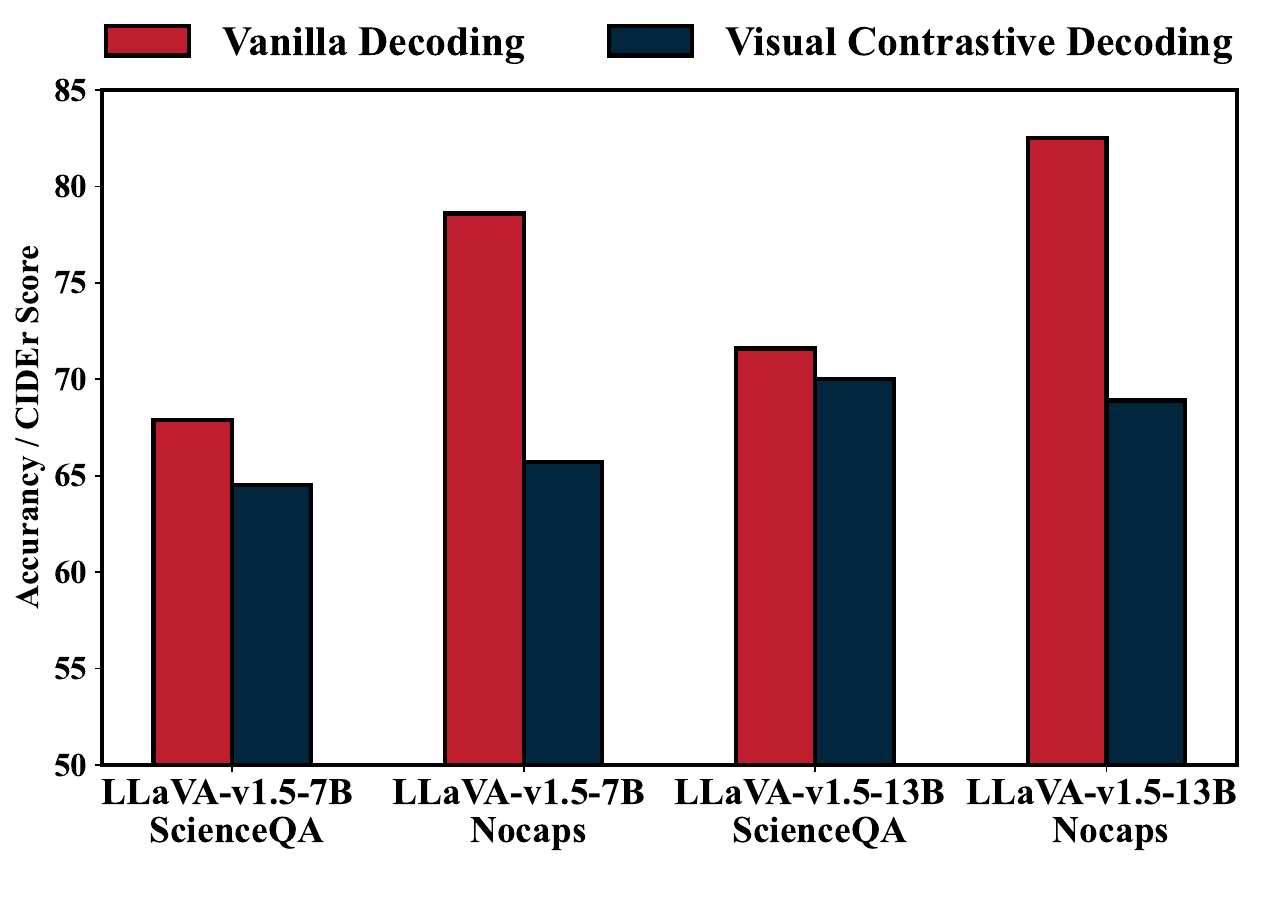}
   \vspace{-0.25cm}
   \caption{\textbf{Impact of VCD on Model Performance.} CIDEr scores are reported on the Nocaps benchmark, while Accuracy is presented for the ScienceQA benchmark. The use of VCD leads to a significant decline in model performance.}
   \label{fig: vcd coherence}
\end{figure}

While contrasting logits of $p_\theta(y \mid v, x)$ and $p_\theta\left(y \mid v^{\prime}, x\right)$ can help reduce over-reliance on language priors and mitigate hallucination in MLLMs-as evidenced by a $4 \%$ performance gain on the POPE benchmark using the VCD method-merely decreasing the influence of the language modality on the output distribution may undermine the coherence of the generated content, potentially leading to prediction errors. This issue is less pronounced in straightforward object hallucination tasks, where responses are limited to binary options, such as "yes" or "no". However, in more complex tasks, including multiple-choice question answering and image caption, the impact of contrastive learning methods on content quality becomes more significant.

To verify this, we applied VCD method to LLaVA-v1.5-7B and LLaVA-v1.5-13B models, assessing their performance on the ScienceQA~\cite{lu2022learn} and NoCaps benchmarks. As illustrated in \cref{fig: vcd coherence}, our findings reveal that, following the application of VCD, model performance decreased by $5 \%$ on ScienceQA and by a considerable $45 \%$ on NoCaps. These results suggest that in tasks requiring nuanced natural language generation, contrastive decoding methods can substantially impair content quality.

\begin{table}[h]
\centering
\begin{tabular}{@{}c|c|cc@{}}
\toprule[1pt]
\toprule
\textbf{Model} & \textbf{Method} & \textbf{ScienceQA} & \textbf{Nocaps} \\ \midrule
               & Regular          & 0.141s             & 0.456s           \\
\multirow{-2}{*}{LLaVA-v1.5-7B}  & VCD & {\color[HTML]{CB0000} \textbf{0.293s}} & {\color[HTML]{CB0000} \textbf{1.086s}} \\ \midrule
               & Regular           & 0.222s              & 0.602s           \\
\multirow{-2}{*}{LLaVA-v1.5-13B} & VCD & {\color[HTML]{CB0000} \textbf{0.459s}} & {\color[HTML]{CB0000} \textbf{1.372s}} \\ \bottomrule
\bottomrule[1pt]
\end{tabular}
\caption{\textbf{Impact of VCD on Model Inference Speed.} The table shows the average inference time per sample (in seconds) on the ScienceQA and Nocaps benchmarks. Applying the VCD method nearly doubled the inference time of the model.}
\label{tab: inference vcd}
\end{table}

Contrastive decoding methods notably reduce inference speed because they require calculating the output distribution for additional contrastive samples. For instance, in VCD method, each visual input $v$ necessitates computing the logits of both $p_\theta(y \mid v, x)$ and $p_\theta\left(y \mid v^{\prime}, x\right)$ separately. This doubles the computation load during inference compared to vanilla decoding. We evaluated the inference speed of VCD versus vanilla decoding on ScienceQA. The experimental results, shown in \cref{tab: inference vcd}, reveal that VCD's inference time is almost double that of vanilla decoding.

%% file: sec/4_visual_neglect.tex
\section{Visual Neglect in Modal Fusion}
\label{sec: visual negelect}

The primary objective of this section is to examine why MLLMs tend to rely excessively on language priors in their predictions. In \cref{subsec: mid-layer}, saliency analysis reveals that image tokens influence prediction outcomes mainly through interactions with instruction tokens within the middle layers. \cref{subsec: attention imbalance} then compares attention weights across different modalities, showing that the attention given to visual features is notably lower than that allocated to system prompts and user instructions. These findings indicate that visual information is often underutilized in the modality fusion process, resulting in an over-reliance on language priors.

\subsection{Mid-layer: Visual-Language Fusion}
\label{subsec: mid-layer}

To uncover why MLLMs tend to overly rely on language priors and overlook visual content in prediction, it is necessary first to clarify how the model utilizes visual modality information. This section explores the influence of the visual modality on prediction outcomes from the perspective of visual information interaction.

We employ the saliency technique, a widely used interpretability tool, to highlight key token interactions within the attention mechanism. Following established practices, we utilize Taylor expansion to compute saliency scores for each element of the attention matrix:
\begin{equation}
    I_l=\left|\sum_h A_{h, l} \odot \frac{\partial \mathcal{L}(x)}{\partial A_{h, l}}\right|.
\end{equation}
Here, $A_{h, l}$ represents the attention matrix value for the $h$-th attention head in the $l$-th layer, $x$ denotes the input, and $\mathcal{L}(x)$ is the loss function of the task, e.g., the cross-entropy objective for question-answering tasks. The saliency matrix $I_l$ for the $l$-th layer is obtained by averaging across all attention heads. The significance of information flow from the $j$-th token to the $i$-th token in MLLMs is represented by $I_l(i, j)$.

To draw a clearer picture of visual information flow in MLLMs, we introduce two quantitative metrics based on $I_l(i, j)$, with a particular focus on the information interaction involving image tokens. The definitions of the two quantitative metrics follow below. \newline
\textbf{$\boldsymbol{S_{vv}}$, measuring the importance of information flow among image tokens:}
\begin{equation}
    \begin{aligned}
    S_{v v} & =\frac{\sum_{(i, j) \in C_{v v}} I_l(i, j)}{\left|C_{v v}\right|} \\
    C_{v v} & =\left\{\left(i, j\right): i, j \in \mathcal{V}, i \geq j\right\} .
    \end{aligned}
\end{equation}
\textbf{$\boldsymbol{S_{vt}}$, measuring the importance of information flow from image tokens to instruction tokens:}
\begin{equation}
    \begin{aligned}
    S_{v t} & =\frac{\sum_{(i, j) \in C_{v t}} I_l(i, j)}{\left|C_{v t}\right|} \\
    C_{v t} & =\left\{\left(i, j\right): i \in \mathcal{T}, j \in \mathcal{V}\right\} .
    \end{aligned}
\end{equation}
Here, $\mathcal{V}$ represents the index set of image tokens, derived from features learned by pre-trained visual encoders, while $\mathcal{T}$ denotes the index set of instruction tokens, specifying requests or questions related to the images. $S_{vv}$ and $S_{vt}$ are utilized to analyze the mechanisms of visual information processing in MLLMs. We define attention interactions among image tokens as \textit{intra-visual information flow} and those between image and instruction tokens as \textit{visual-textual information flow}. 

\begin{figure}[h]
  \centering
   \includegraphics[width=0.9\linewidth]{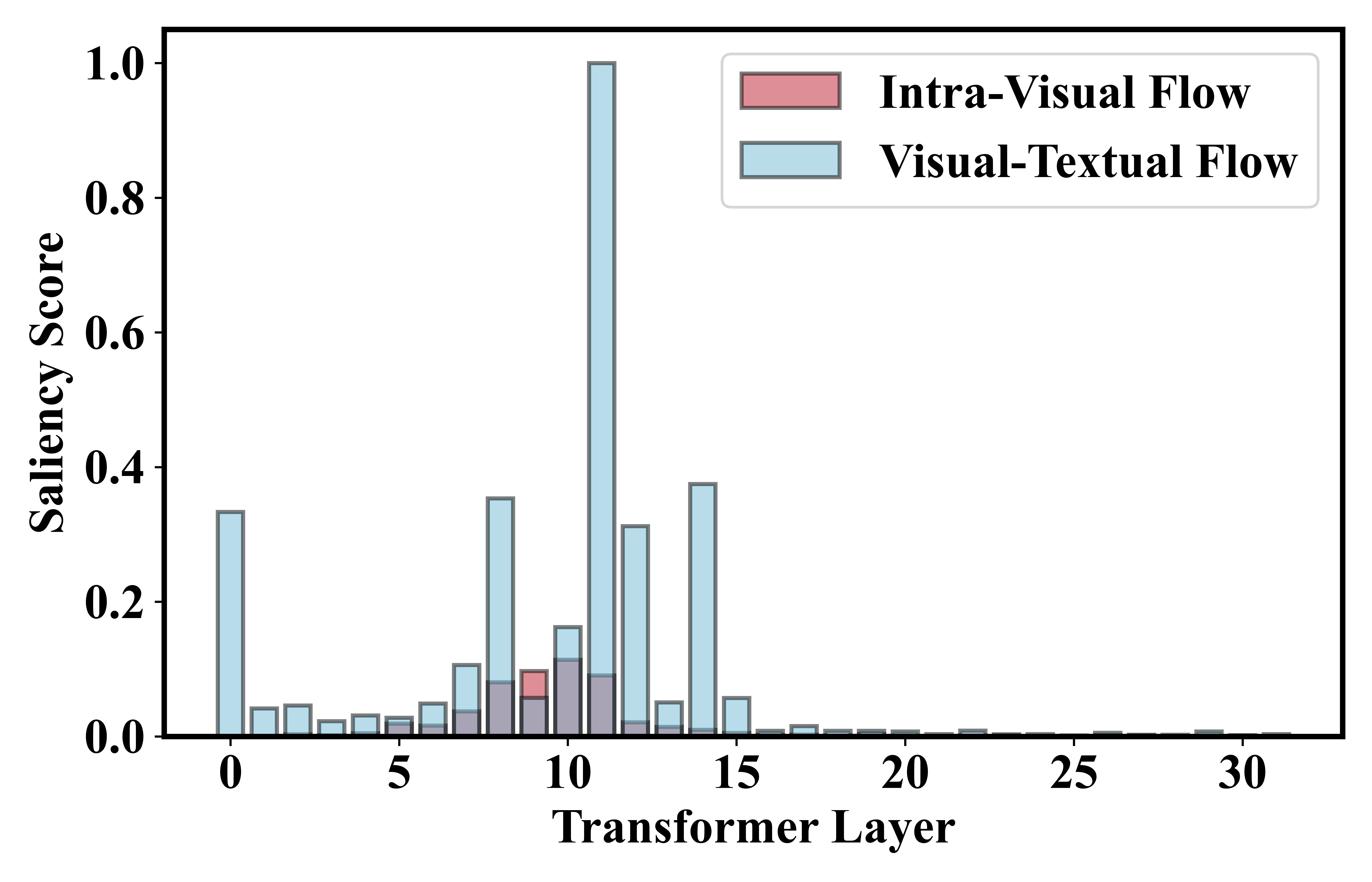}
   \vspace{-0.25cm}
   \caption{\textbf{The importance of \textit{intra-visual flow} and \textit{visual-textual flow} across various layers.} The\textit{ visual-textual information flow} in the middle layers has a significant impact on prediction outcomes.}
   \label{fig: mid fusion}
\end{figure}

We conducted experiments with the LLaVA-v1.5-7B model on the MS COCO dataset under the POPE benchmark, sampling 500 examples for evaluation. \cref{fig: mid fusion} underscores the critical role of the visual-textual information flow within the model’s middle layers, specifically from the 8-th to the 15-th layer. This observation indicates that in these layers, visual information interacts intensively with textual information via attention mechanisms, which substantially influences the prediction outcomes. 

\vspace{-0.35cm}

\subsection{Attention Imbalance Across Modalities}
\label{subsec: attention imbalance}

\cref{subsec: mid-layer} reveals that the middle layers facilitate crucial fusion, integrating visual and textual inputs into cross-modal semantic representations that drive final predictions. Accordingly, this section will delve deeper into the attention to visual inputs throughout the modality fusion process.

\begin{figure}[h]
  \centering
   \includegraphics[width=0.95\linewidth]{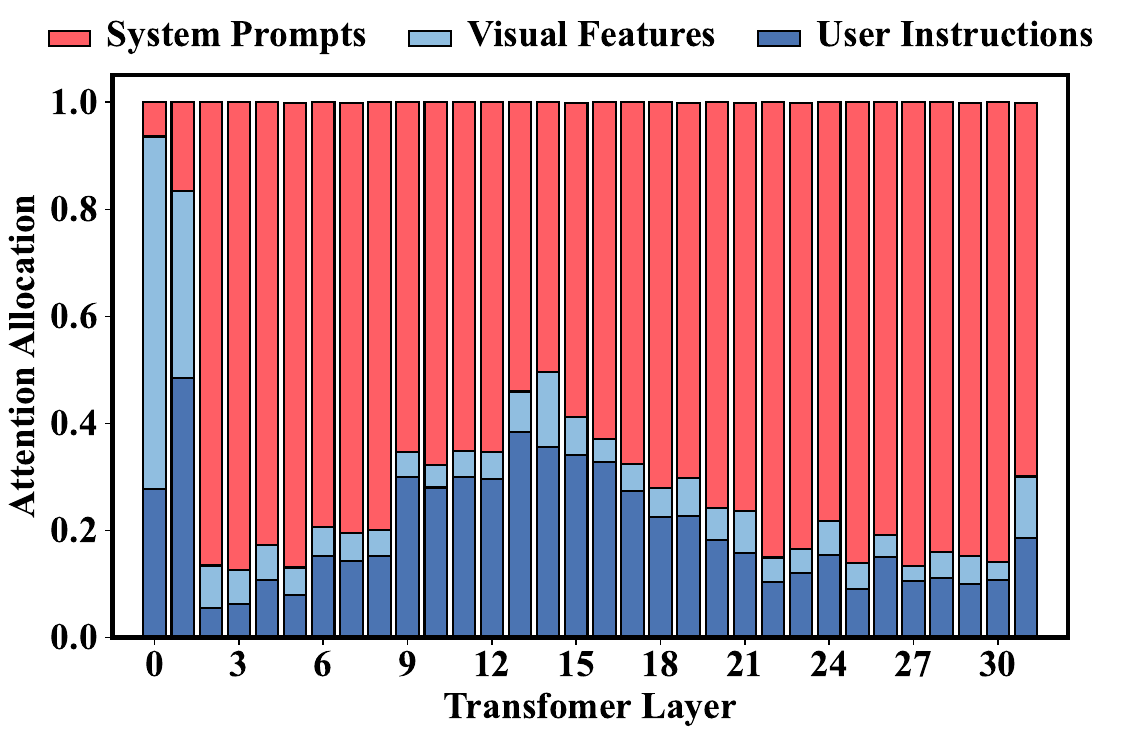}
   \vspace{-0.25cm}
   \caption{\textbf{Attention Distribution of Modal Information Across Model Layers.} In the middle layers, the model allocates insufficient attention to visual features while disproportionately focusing on system prompts.}
   \label{fig: mid attention}
\end{figure}

We define the attention allocation, $\lambda$, as the aggregate attention score assigned to a specific type of token within a single layer. Accordingly, the attention allocation for system prompts, visual features, and user instructions in the $l$-th layer can be computed as follows:

\begin{equation}
    \begin{aligned}
        \lambda_{sys}^l &= \sum_{i \in \mathcal{T}} \sum_{j \in \mathcal{S}} A_l(i,j), \\
        \lambda_{vis}^l &= \sum_{i \in \mathcal{T}} \sum_{j \in \mathcal{V}} A_l(i,j), \\
        \lambda_{ins}^l &= \sum_{i \in \mathcal{T}} \sum_{j \in \mathcal{T}} A_l(i,j). \\
    \end{aligned}
\end{equation}
In this context, $A_l$ represents the attention matrix averaged across all attention heads, while $\mathcal{S}$ represents the indices of system tokens. The measures $\lambda_{sys}^l, \lambda_{vis}^l$, and $\lambda_{ins}^l$ provide insight into the distribution of attention to different modalities across various layers, aiding in understanding the reasons for the underutilization of visual information during the modality fusion process.

The experimental setup aligns with that described in \cref{subsec: mid-layer}. \cref{fig: mid attention} illustrates the allocation of attention to different modalities across the model's layers. In the middle layers, attention to visual features is markedly lower than that given to system prompts and user instructions. This suggests that during the critical process of modality fusion, the model's focus on visual input is insufficient. As a result, visual information is underutilized, leading to an output distribution skewed toward language priors.

\subsection{Insights}
Based on the experimental results presented in \cref{subsec: mid-layer} and \cref{subsec: attention imbalance}, two significant conclusions can be drawn:

\begin{itemize}
    \item The model performs the crucial fusion of visual and textual modalities in the middle layers, creating cross-modal semantic representations that drive the final predictions.
    \item During this critical fusion process, the model demonstrates inadequate attention to the visual modality.
\end{itemize}
These findings indicate that models fail to fully utilize visual information, resulting in an excessive dependence on language priors and, subsequently, the occurrence of hallucination phenomena.

%% file: sec/5_method.tex
\begin{figure}[t]
  \centering
   \includegraphics[width=0.9\linewidth]{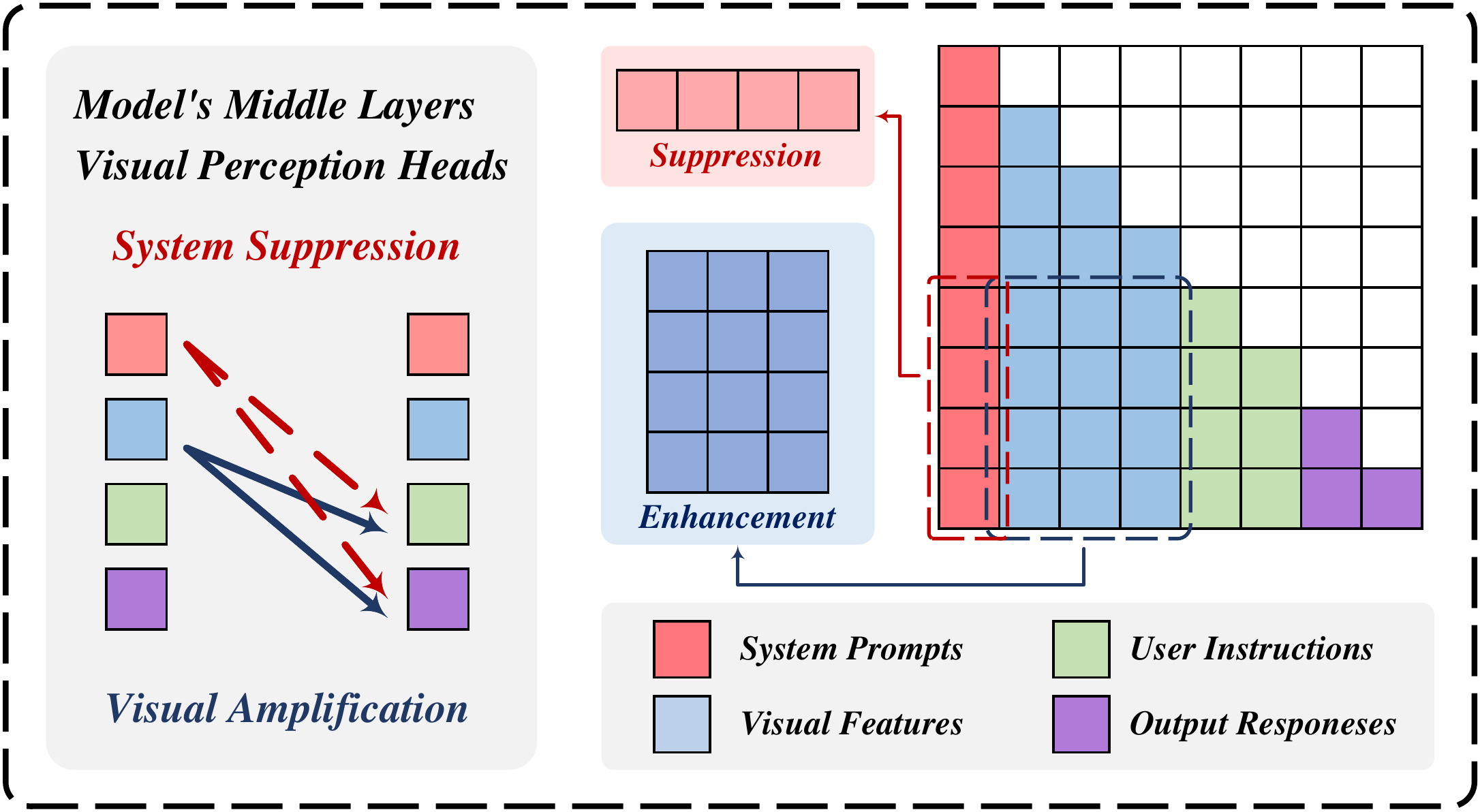}
   \vspace{-0.15cm}
   \caption{\textbf{Illustration of the Visual Amplification Fusion Method.} In the middle layers, we select attention heads highly responsive to visual information, amplifying their focus on visual features while reducing unnecessary attention to system prompts.}
   \label{fig: VAF}
\end{figure}

\section{Visual Amplification Fusion}
 Building on the insights presented in \cref{sec: visual negelect}, we introduce a hallucination mitigation method called Visual Amplification Fusion (VAF). As illustrated in \cref{fig: VAF}, This approach heightens attention to visual information during modality fusion, effectively reducing the excessive dependency on language priors and ensuring that the generated content is closely grounded to visual inputs.
 
\subsection{Attention Redistribution}

As outlined in \cref{sec: visual negelect}, the model performs crucial fusion of visual and textual modalities within the middle layers. However, the attention allocated to visual modality information during this process remains insufficient. To address this, we adjust the attention weights in these layers to achieve a more balanced focus.

\begin{figure}[h]
  \centering
   \includegraphics[width=0.9\linewidth]{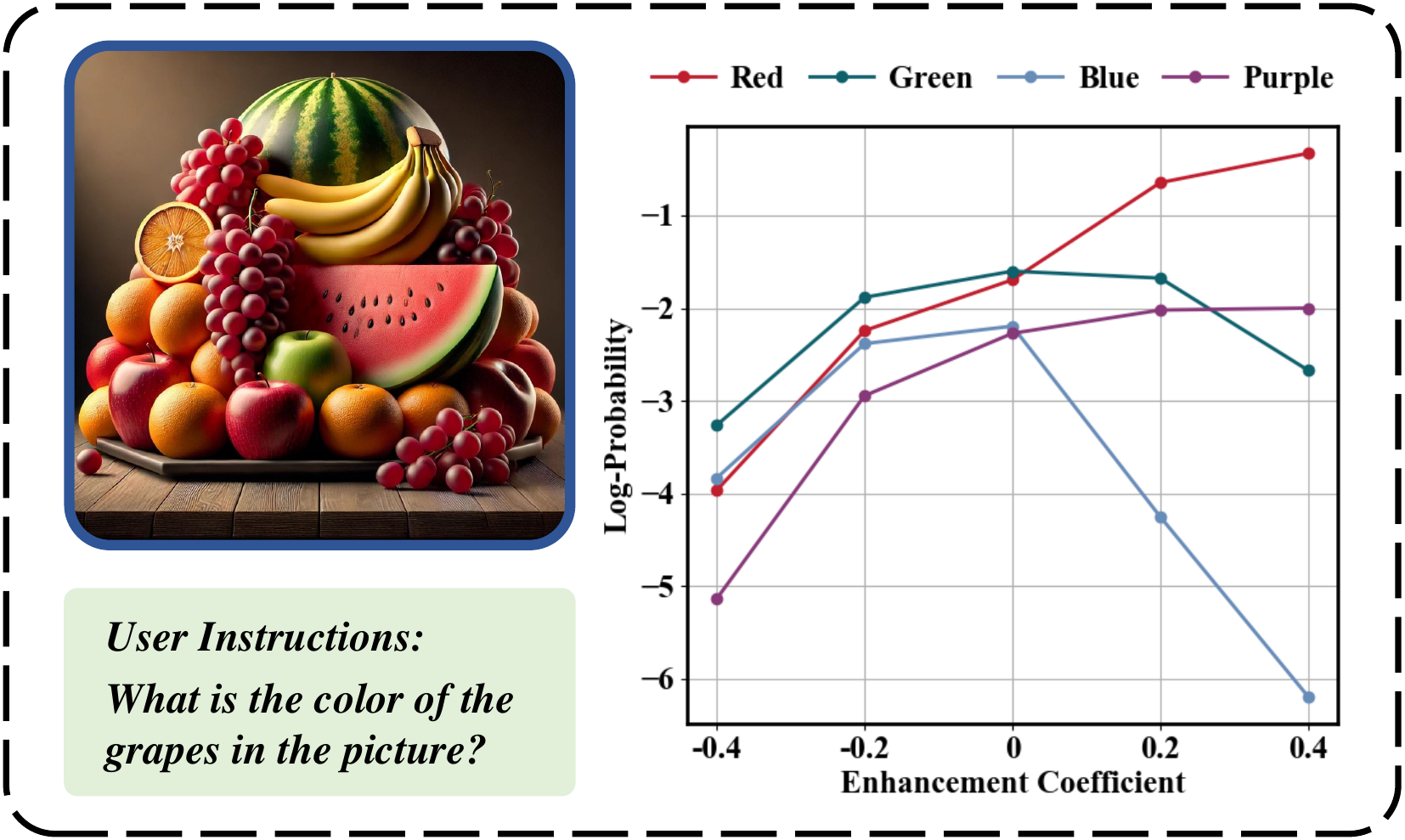}
   \caption{\textbf{Effect of Enhanced Visual Attention on Hallucination Suppression.} Increasing attention to visual features in the fusion process of the model's middle layers successfully reduces hallucinations, enabling the model to correct its grape color prediction from "green" to "red".}
   \label{fig: Probs_VAF}
\end{figure}

Let $A_{l, h}$ denote the attention matrix of the $h$-th attention head in the $l$-th layer, and $Z_{l, h}$ represent its corresponding attention score matrix, defined as:
\begin{equation}
    A_{l, h}=\operatorname{softmax}\left(Z_{l, h}\right).
\end{equation}

Our objective during the modality fusion process is to amplify the model's attention to visual features while curbing an overemphasis on system prompts. This adjustment facilitates improved integration of visual information and reduces over-reliance on language priors. To achieve this, we modify the attention score matrix in the middle layers (i.e., $8 <l<15$) as follows:
\begin{equation}
    \hat{Z}_{l,h} = Z_{l,h} + \alpha \cdot M^{enh}_{l,h} \circ Z_{l,h} - \beta \cdot M^{sup}_{l,h} \circ Z_{l,h}.
\end{equation}
Here, $\alpha$ is the enhancement coefficient ($\alpha>0$), where larger values indicate stronger amplification of visual attention. The suppression coefficient $\beta$ ($0<\beta<1$) determines the extent of attention suppression directed at system prompts. The enhancement and suppression mask matrices, $M_{l, h}^{e n h}$ and $M_{l, h}^{s u p}$ respectively, are defined to guide the modulation of attention elements:
\begin{equation}
\begin{aligned}
    M_{l, h}^{enh}(i, j) &= \mathbb{I}(i \in \mathcal{T}, j \in \mathcal{V}), \\
    M_{l, h}^{sup}(i, j) &= \mathbb{I}(i \in \mathcal{T}, j \in \mathcal{S}).
\end{aligned}
\end{equation}

These modifications optimize attention allocation by enhancing the model's focus on visual features during modality fusion and minimizing superfluous attention to system prompts. As illustrated in \cref{fig: Probs_VAF}, preliminary analysis indicates that this approach effectively mitigates hallucination issues by promoting greater attention to visual information.

\begin{table*}[b]
\centering
\begin{tabular}{@{}cc|cc|cc|cc@{}}
\toprule[1pt]
\toprule
 &
   &
  \multicolumn{2}{c|}{\textbf{LLaVA-v1.5-7B}} &
  \multicolumn{2}{c|}{\textbf{LLaVA-v1.5-13B}} &
  \multicolumn{2}{c}{\textbf{Qwen-VL-Chat-7B}} \\ \cmidrule(l){3-8} 
\multirow{-2}{*}{\textbf{Category}} &
  \multirow{-2}{*}{\textbf{Method}} &
  Accuracy &
  F1-score &
  Accuracy &
  F1-score &
  Accuracy &
  F1-score \\ \midrule
 &
  Regular &
  87.8 \(\uparrow 0.0\) &
  87.5 \(\uparrow 0.0\) &
  87.6 \(\uparrow 0.0\) &
  87.4 \(\uparrow 0.0\) &
  88.2 \(\uparrow 0.0\) &
  87.9 \(\uparrow 0.0\) \\
 &
  VCD &
  88.4 \(\uparrow 0.6\) &
  87.7 \(\uparrow 0.2\) &
  88.9 \(\uparrow 1.3\) &
  87.8 \(\uparrow 0.4\) &
  89.1 \(\uparrow 0.9\) &
  88.4 \(\uparrow 0.5\) \\
 &
  ICD &
  88.1 \(\uparrow 0.3\) &
  87.6 \(\uparrow 0.1\) &
  88.1 \(\uparrow 0.5\) &
  87.6 \(\uparrow 0.2\) &
  88.9 \(\uparrow 0.7\) &
  88.1 \(\uparrow 0.2\) \\
\multirow{-4}{*}{Random} &
  \cellcolor[HTML]{EFEFEF}VAF &
  \cellcolor[HTML]{EFEFEF}{\color[HTML]{CB0000} \textbf{89.6} \(\uparrow 1.8\)} &
  \cellcolor[HTML]{EFEFEF}{\color[HTML]{CB0000} \textbf{89.3} \(\uparrow 1.8\)} &
  \cellcolor[HTML]{EFEFEF}{\color[HTML]{CB0000} \textbf{90.1} \(\uparrow 2.5\)} &
  \cellcolor[HTML]{EFEFEF}{\color[HTML]{CB0000} \textbf{89.9} \(\uparrow 2.5\)} &
  \cellcolor[HTML]{EFEFEF}{\color[HTML]{CB0000} \textbf{90.0} \(\uparrow 1.8\)} &
  \cellcolor[HTML]{EFEFEF}{\color[HTML]{CB0000} \textbf{89.7} \(\uparrow 1.8\)} \\ \midrule
 &
  Regular &
  82.5 \(\uparrow 0.0\) &
  83.2 \(\uparrow 0.0\) &
  82.7 \(\uparrow 0.0\) &
  84.1 \(\uparrow 0.0\) &
  82.4 \(\uparrow 0.0\) &
  83.1 \(\uparrow 0.0\) \\
 &
  VCD &
  83.1 \(\uparrow 0.6\) &
  84.1 \(\uparrow 0.9\) &
  83.7 \(\uparrow 1.0\) &
  85.1 \(\uparrow 1.0\) &
  83.0 \(\uparrow 0.6\) &
  84.1 \(\uparrow 1.0\) \\
 &
  ICD &
  82.1 \(\downarrow 0.4\) &
  82.9 \(\downarrow 0.3\) &
  82.9 \(\uparrow 0.2\) &
  84.3 \(\uparrow 0.2\) &
  83.2 \(\uparrow 0.8\) &
  84.5 \(\uparrow 1.4\) \\
\multirow{-4}{*}{Popular} &
  \cellcolor[HTML]{EFEFEF}VAF &
  \cellcolor[HTML]{EFEFEF}{\color[HTML]{CB0000} \textbf{84.5} \(\uparrow 2.0\)} &
  \cellcolor[HTML]{EFEFEF}{\color[HTML]{CB0000} \textbf{84.9} \(\uparrow 1.7\)} &
  \cellcolor[HTML]{EFEFEF}{\color[HTML]{CB0000} \textbf{85.2} \(\uparrow 2.5\)} &
  \cellcolor[HTML]{EFEFEF}{\color[HTML]{CB0000} \textbf{86.4} \(\uparrow 2.3\)} &
  \cellcolor[HTML]{EFEFEF}{\color[HTML]{CB0000} \textbf{84.9} \(\uparrow 2.5\)} &
  \cellcolor[HTML]{EFEFEF}{\color[HTML]{CB0000} \textbf{85.1} \(\uparrow 2.0\)} \\ \midrule
 &
  Regular &
  77.6 \(\uparrow 0.0\) &
  79.4 \(\uparrow 0.0\) &
  77.8 \(\uparrow 0.0\) &
  79.5 \(\uparrow 0.0\) &
  77.2 \(\uparrow 0.0\) &
  78.9 \(\uparrow 0.0\) \\
 &
  VCD &
  78.1 \(\uparrow 0.5\) &
  79.6 \(\uparrow 0.2\) &
  78.2 \(\uparrow 0.4\) &
  79.7 \(\uparrow 0.2\) &
  78.8 \(\uparrow 1.6\) &
  80.1 \(\uparrow 1.2\) \\
 &
  ICD &
  78.5 \(\uparrow 0.9\) &
  79.9 \(\uparrow 0.5\) &
  79.1 \(\uparrow 1.3\) &
  80.1 \(\uparrow 0.6\) &
  78.1 \(\uparrow 0.9\) &
  79.2 \(\uparrow 0.3\) \\
\multirow{-4}{*}{Adversarial} &
  \cellcolor[HTML]{EFEFEF}VAF &
  \cellcolor[HTML]{EFEFEF}{\color[HTML]{CB0000} \textbf{80.1} \(\uparrow 2.5\)} &
  \cellcolor[HTML]{EFEFEF}{\color[HTML]{CB0000} \textbf{81.0} \(\uparrow 1.6\)} &
  \cellcolor[HTML]{EFEFEF}{\color[HTML]{CB0000} \textbf{80.7} \(\uparrow 2.9\)} &
  \cellcolor[HTML]{EFEFEF}{\color[HTML]{CB0000} \textbf{81.7} \(\uparrow 2.2\)} &
  \cellcolor[HTML]{EFEFEF}{\color[HTML]{CB0000} \textbf{80.4} \(\uparrow 3.2\)} &
  \cellcolor[HTML]{EFEFEF}{\color[HTML]{CB0000} \textbf{81.2} \(\uparrow 2.3\)} \\ \bottomrule
\bottomrule[1pt]
\end{tabular}
\caption{\textbf{Performance on POPE.} Results are averaged across the MS-COCO, A-OKVQA, and GQA datasets. The VAF method demonstrates superior hallucination suppression across all three MLLMs. The {\color[HTML]{CB0000} \textbf{best performance}} for each setting is highlighted in {\color[HTML]{CB0000} \textbf{red}}.}
\label{tab: pope}
\end{table*}

\begin{table*}[t]
\centering
\begin{tabular}{@{}c|c|cc|cc|c@{}}
\toprule[1pt]
\toprule
 &
   &
  \multicolumn{2}{c|}{\textbf{Object-level}} &
  \multicolumn{2}{c|}{\textbf{Attribute-level}} &
   \\ \cmidrule(lr){3-6}
\multirow{-2}{*}{\textbf{Model}} &
  \multirow{-2}{*}{\textbf{Method}} &
  Existence &
  Count &
  Position &
  Color &
  \multirow{-2}{*}{\textbf{Total Score}} \\ \midrule
 &
  Regular &
  185.00 \(\uparrow 0.00\) &
  146.67 \(\uparrow 0.00\) &
  128.33 \(\uparrow 0.00\) &
  150.00 \(\uparrow 0.00\) &
  610.00 \(\uparrow 0.00\) \\
 &
  VCD &
  185.00 \(\uparrow 0.00\) &
  141.33 \(\downarrow 5.34\) &
  128.33 \(\uparrow 0.00\) &
  153.00 \(\uparrow 3.00\) &
  607.66 \(\downarrow 2.34\) \\
 &
  ICD &
  185.00 \(\uparrow 0.00\) &
  148.33 \(\uparrow 1.66\) &
  126.66 \(\downarrow 1.67\) &
  148.33 \(\downarrow 1.67\) &
  608.32 \(\downarrow 1.68\) \\
\multirow{-4}{*}{LLaVA-v1.5-7B} &
  \cellcolor[HTML]{EFEFEF}VAF &
  \cellcolor[HTML]{EFEFEF}{\color[HTML]{CB0000} \textbf{195.00} \(\uparrow 10.00\)} &
  \cellcolor[HTML]{EFEFEF}{\color[HTML]{CB0000} \textbf{158.33} \(\uparrow 11.66\)} &
  \cellcolor[HTML]{EFEFEF}{\color[HTML]{CB0000} \textbf{128.33} \(\uparrow 0.00\)} &
  \cellcolor[HTML]{EFEFEF}{\color[HTML]{CB0000} \textbf{155.00} \(\uparrow 5.00\)} &
  \cellcolor[HTML]{EFEFEF}{\color[HTML]{CB0000} \textbf{636.67} \(\uparrow 26.67\)} \\ \midrule
 &
  Regular &
  185.00 \(\uparrow 0.00\) &
  155.00 \(\uparrow 0.00\) &
  133.33 \(\uparrow 0.00\) &
  165.00 \(\uparrow 0.00\) &
  638.33 \(\uparrow 0.00\) \\
 &
  VCD &
  185.00 \(\uparrow 0.00\) &
  155.00 \(\uparrow 0.00\) &
  130.00 \(\downarrow 3.33\) &
  168.33 \(\uparrow 3.33\) &
  638.33 \(\uparrow 0.00\) \\
 &
  ICD &
  183.33 \(\downarrow 1.67\) &
  153.33 \(\downarrow 1.67\) &
  131.67 \(\downarrow 1.66\) &
  165.00 \(\uparrow 0.00\) &
  633.33 \(\downarrow 5.00\) \\
\multirow{-4}{*}{LLaVA-v1.5-13B} &
  \cellcolor[HTML]{EFEFEF}VAF &
  \cellcolor[HTML]{EFEFEF}{\color[HTML]{CB0000} \textbf{195.00} \(\uparrow 10.00\)} &
  \cellcolor[HTML]{EFEFEF}{\color[HTML]{CB0000} \textbf{160.00} \(\uparrow 5.00\)} &
  \cellcolor[HTML]{EFEFEF}{\color[HTML]{CB0000} \textbf{136.67} \(\uparrow 3.34\)} &
  \cellcolor[HTML]{EFEFEF}{\color[HTML]{CB0000} \textbf{170.00} \(\uparrow 5.00\)} &
  \cellcolor[HTML]{EFEFEF}{\color[HTML]{CB0000} \textbf{661.67} \(\uparrow 23.34\)} \\ \midrule
 &
  Regular &
  158.33 \(\uparrow 0.00\) &
  150.00 \(\uparrow 0.00\) &
  128.33 \(\uparrow 0.00\) &
  170.00 \(\uparrow 0.00\) &
  606.66 \(\uparrow 0.00\) \\
 &
  VCD &
  158.33 \(\uparrow 0.00\) &
  150.00 \(\uparrow 0.00\) &
  133.33 \(\uparrow 5.00\) &
  175.00 \(\uparrow 5.00\) &
  616.66 \(\uparrow 10.00\) \\
 &
  ICD &
  128.33 \(\downarrow 30.00\) &
  151.67 \(\uparrow 1.67\) &
  128.33 \(\uparrow 0.00\) &
  170.00 \(\uparrow 0.00\) &
  578.33 \(\downarrow 28.33\) \\
\multirow{-4}{*}{Qwen-VL-7B} &
  \cellcolor[HTML]{EFEFEF}VAF &
  \cellcolor[HTML]{EFEFEF}{\color[HTML]{CB0000} \textbf{165.00} \(\uparrow 6.67\)} &
  \cellcolor[HTML]{EFEFEF}{\color[HTML]{CB0000} \textbf{155.00} \(\uparrow 5.00\)} &
  \cellcolor[HTML]{EFEFEF}{\color[HTML]{CB0000} \textbf{133.33} \(\uparrow 5.00\)} &
  \cellcolor[HTML]{EFEFEF}{\color[HTML]{CB0000} \textbf{175.00} \(\uparrow 5.00\)} &
  \cellcolor[HTML]{EFEFEF}{\color[HTML]{CB0000} \textbf{628.33} \(\uparrow 21.67\)} \\ \bottomrule
\bottomrule[1pt]
\end{tabular}
\caption{\textbf{Results on the MME subset.} Across three MLLMs, the VAF method achieved the most effective suppression of both object-level and attribute-level hallucinations. The {\color[HTML]{CB0000} \textbf{highest scores}} in each setting are highlighted in {\color[HTML]{CB0000} \textbf{red}}.}
\label{tab: mme subset}
\end{table*}

\subsection{Visual Perception Restriction}
Enhancing visual attention across all attention heads in the middle layers can be overly aggressive and may negatively impact content generation. To address this, we propose a selective enhancement strategy. Specifically, we identify and isolate the attention heads that exhibit higher sensitivity to visual information, which we term\textit{ visual perception heads}. We then restrict the visual attention enhancement to these\textit{ visual perception heads}, ensuring better utilization of visual information while maintaining overall model performance.

In the model, attention heads that allocate more attention to visual features demonstrate heightened sensitivity to visual information. Let $A_{l, h}$ represent the attention matrix of the $h$-th attention head in the $l$-th layer of the model, with its corresponding visual attention allocation denoted by $\lambda_{\text {vis }}^{l, h}$. In each attention layer, we identify the attention heads whose visual attention allocation fall within the top $50 \%$ and designate them as \textit{visual perception heads}, subsequently redistributing their attention. The attention matrices of the remaining attention heads are kept unchanged.


%% file: sec/6_experiment.tex
\section{Experiment}
This section demonstrates the effectiveness of the proposed VAF method in mitigating hallucinations. \cref{subsec: experimental settings} outlines the experimental setup, detailing the evaluation benchmarks and VAF parameter configurations. \cref{subsec: results} then presents the experimental results from three perspectives: reduction of hallucinations, coherence of generated content, and inference speed. Finally, \cref{subsec: ablation study} further verifies the contribution of each VAF component through ablation studies.

\subsection{Experimental Settings}
\label{subsec: experimental settings}

In \cref{subsubsec: datasets}, we present the selected datasets and evaluation metrics. \cref{subsubsec: mllm backbones} details the chosen MLLM backbone models, and \cref{subsubsec: baseline settings.} outlines the baseline settings.

\subsubsection{Datasets \& Evaluation Metrics}
\label{subsubsec: datasets}

\textbf{Polling-based Object Probing Evaluation (POPE).} POPE~\cite{pope} is a novel framework designed to evaluate object hallucinations in MLLMs. Departing from traditional caption-based approaches, POPE frames hallucination detection as a binary task by posing straightforward yes-or-no questions regarding the presence of specific objects in an image (e.g., "Is there a chair in the image?"). Performance on POPE is measured across four metrics: Accuracy, Precision, Recall, and F1 score, allowing for a thorough evaluation of hallucinations in MLLMs.

\noindent \textbf{Multimodal Model Evaluation (MME).} MME~\cite{mme} benchmark provides a comprehensive framework for evaluating MLLMs across both perceptual and cognitive dimensions. It consists of ten perception-oriented tasks and four cognition-oriented tasks, with model performance assessed through accuracy metrics. In addition to the full dataset, we leverage specific subsets, such as object existence and counting to analyze object-level hallucinations, while position and color subsets are employed to examine attribute-level hallucinations.

\noindent \textbf{Novel Object Captioning at Scale (Nocaps).} NoCaps~\cite{Agrawal_2019} benchmark is designed to evaluate image captioning models on their ability to describe novel objects absent from standard datasets like COCO. Model performance is quantified using the CIDEr score, providing a basis to assess the coherence and accuracy of generated captions in response to images containing unfamiliar objects.

\subsubsection{MLLM Backbones}
\label{subsubsec: mllm backbones}

In comparison to the Q-former structure, linear projection demonstrates greater efficiency in aligning visual and textual features. This advantage is evident in MLLMs with linear projection architectures, such as LLaVA and Qwen-VL, which outperform Q-former-based MLLMs like InstructBLIP and MiniGPT4. Based on these findings, we selected three linear-projection-based MLLMs, specifically LLaVA-v1.5-7B, LLaVA-v1.5-13B~\cite{llava1.5}, and Qwen-VL-7B~\cite{qwenvl}, to evaluate the effectiveness of our proposed VAF method. Detailed prompt templates for each model across various benchmarks are included in \cref{app: prompts}. 

\subsubsection{Baseline Settings.}
\label{subsubsec: baseline settings.}

We primarily compared our approach to the VCD~\cite{vcd} and ICD~\cite{icd} methods. VCD mitigates hallucinations by contrasting output distributions derived from original and distorted visual inputs, while ICD reduces hallucinated concepts by comparing distributions generated with standard versus disrupted instructions. To ensure consistency and reproducibility in our comparisons, all methods use greedy search. Unless specified otherwise, our experiments set $\beta = 0.1$ and $\alpha = 0.15$.

\subsection{Results and Analysis}
\label{subsec: results}

\cref{subsubsec: hallucination} examines the effectiveness of various methods in mitigating hallucinations, while \cref{subsubsec: coherence} assesses their impact on the quality of generated content. \cref{subsubsec: inference} then analyzes the influence of each method on inference speed. Additional experimental results are provided in \cref{app: experimental results}.

\subsubsection{Hallucination Mitigation}
\label{subsubsec: hallucination}

\cref{tab: pope} presents the experimental results of the VAF method on the POPE benchmark, with results averaged across the MSCOCO~\cite{mscoco}, A-OKVQA~\cite{aokvqa}, and GQA~\cite{gqa} datasets. Applied to both the LLaVA-v1.5 model family and the Qwen-VL model, the VAF method consistently surpasses the VCD and ICD methods in reducing hallucinations. \cref{tab: mme subset} further highlights the performance of VAF on the MME benchmark, demonstrating its effectiveness in suppressing both object-level and attribute-level hallucinations.

\begin{table}[t]
\centering
\begin{tabular}{@{}c|c|cc@{}}
\toprule[1pt]
\toprule
                                 &                                     & \textbf{ScienceQA}                   & \textbf{Nocaps}                      \\ \cmidrule(l){3-4} 
\multirow{-2}{*}{\textbf{Model}} & \multirow{-2}{*}{\textbf{Decoding}} & Accurancy                            & CIDEr                                \\ \midrule
                                 & Regular                             & {\color[HTML]{010066} \textbf{68.0}} & {\color[HTML]{010066} \textbf{78.7}} \\
                                 & VCD                                 & 64.5                                 & 65.7                                 \\
                                 & ICD                                 & 62.4                                 & 62.3                                 \\
\multirow{-4}{*}{LLaVA-v1.5-7B} &
  \cellcolor[HTML]{EFEFEF}VAF &
  \cellcolor[HTML]{EFEFEF}{\color[HTML]{CB0000} \textbf{68.5}} &
  \cellcolor[HTML]{EFEFEF}{\color[HTML]{CB0000} \textbf{78.8}} \\ \midrule
                                 & Regular                             & {\color[HTML]{010066} \textbf{71.6}} & {\color[HTML]{CB0000} \textbf{82.6}} \\
                                 & VCD                                 & 70.0                                 & 68.9                                 \\
                                 & ICD                                 & 69.2                                 & 60.3                                 \\
\multirow{-4}{*}{LLaVA-v1.5-13B} &
  \cellcolor[HTML]{EFEFEF}VAF &
  \cellcolor[HTML]{EFEFEF}{\color[HTML]{CB0000} \textbf{71.7}} &
  \cellcolor[HTML]{EFEFEF}{\color[HTML]{010066} \textbf{82.3}} \\ \bottomrule
\bottomrule[1pt]
\end{tabular}
\caption{\textbf{Results on SQA and Nocaps datasets.} The {\color[HTML]{CB0000} \textbf{highest}} and {\color[HTML]{010066} \textbf{second-highest}} scores are marked in {\color[HTML]{CB0000} \textbf{red}} and {\color[HTML]{010066} \textbf{blue}}, respectively.}
\label{tab: sqa & nocaps}
\end{table}

\subsubsection{Coherence of Generated Content}
\label{subsubsec: coherence}

\cref{tab: sqa & nocaps} presents the experimental results for various methods on the Nocaps and ScienceQA datasets. It is evident that VCD and ICD substantially degrade the quality of the generated content. Specifically, on the Nocaps dataset, VCD and ICD reduce CIDEr scores by 18\% and 27\%, respectively. This degradation primarily arises from the crude disruption of language priors by contrastive decoding methods, which leads to generated content lacking coherence and accuracy. By contrast, our method demonstrates minimal negative impact on prediction results, maintaining both coherence and accuracy effectively.

\subsubsection{Inference Speed}
\label{subsubsec: inference}

\begin{figure}[h]
  \centering
   \includegraphics[width=0.9\linewidth]{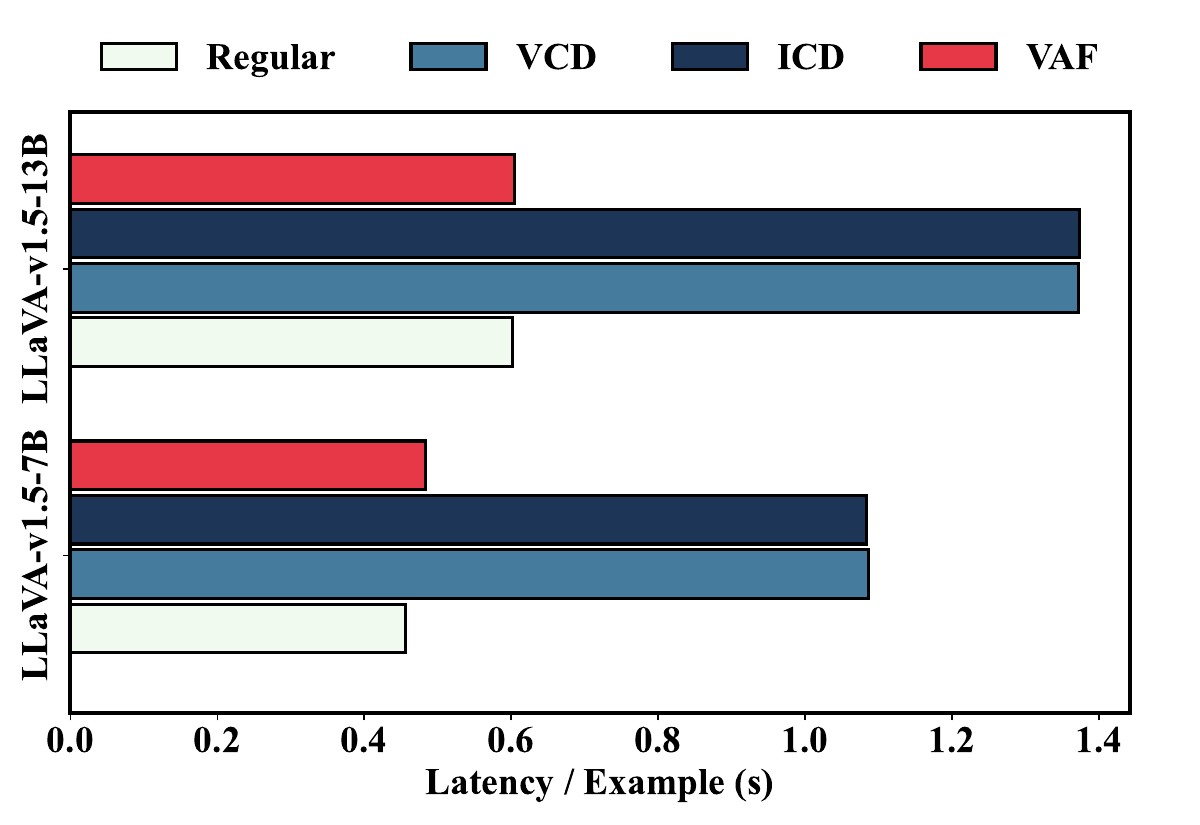}
   \vspace{-0.15cm}
   \caption{\textbf{Comparison of different strategies on inference speed.} The VCD and ICD methods reduce inference speed by 50\%, whereas the VAF method shows minimal impact.}
   \label{fig: inference speed}
\end{figure}

\cref{fig: inference speed} illustrates the impact of different strategies on model inference speed within the Nocaps dataset. In comparison, the VCD and ICD methods nearly double the inference time due to the need to process contrastive input samples, whereas the VAF method has minimal impact on the inference speed of multimodal large language models.

\vspace{-0.1cm}

\subsection{Ablation Study}
\label{subsec: ablation study}

\begin{figure}[h]
  \centering
   \includegraphics[width=0.9\linewidth]{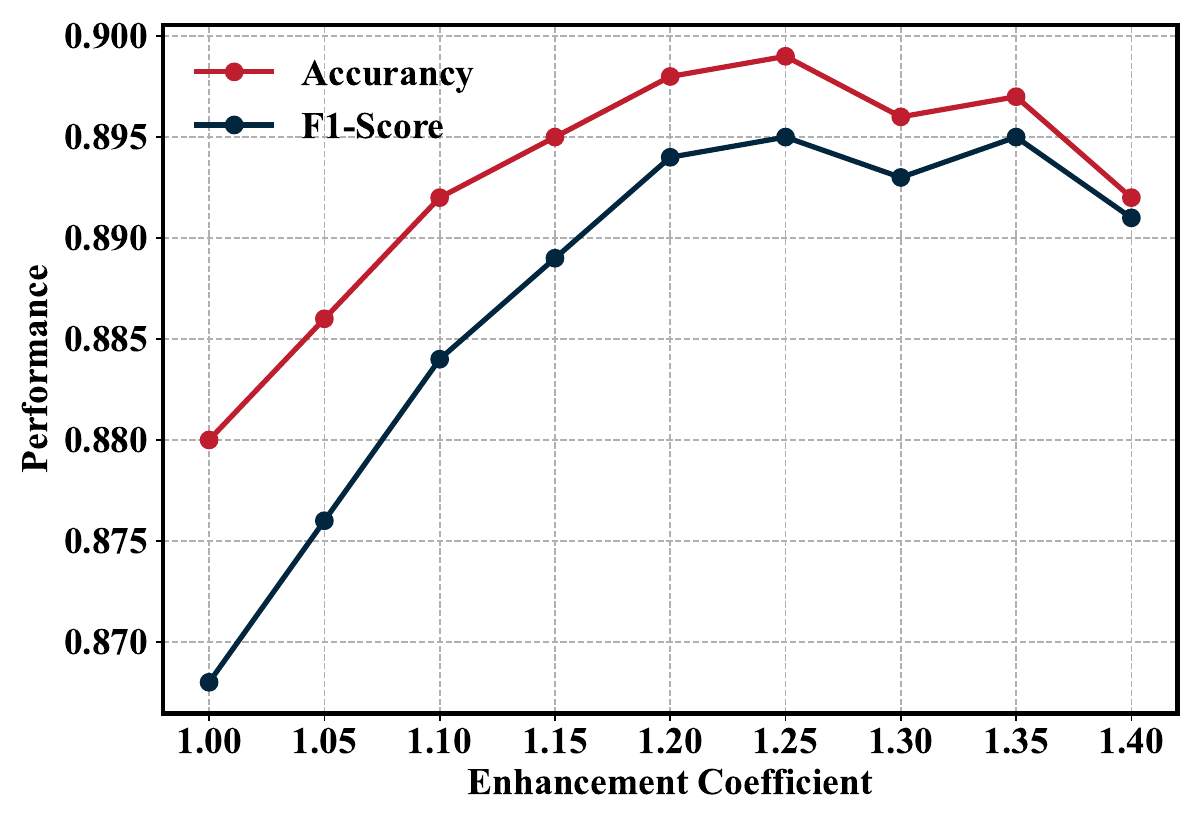}
   \vspace{-0.15cm}
   \caption{\textbf{Ablation study of $\alpha$ on the POPE benchmark.}}
   \label{fig: ablation alpha}
\end{figure}

Ablation studies on the enhancement coefficient $\alpha$ were conducted using the COCO-Random dataset within the POPE benchmark to understand its influence on model performance. \cref{fig: ablation alpha} demonstrates that when $0 < \alpha < 0.25$, model hallucinations are effectively suppressed. However, when $\alpha$ surpasses 0.25, performance starts to degrade. We propose that this reduction in performance may stem from an excessive focus on visual features, disrupting the balanced integration of language information and diminishing overall model effectiveness.

\vspace{-0.35cm}

We performed ablation studies on the visual perception restriction mechanism, evaluating its impact on the POPE and Nocaps benchmarks. \cref{tab: visual restriction} highlights the effects of restricting attention reallocation to \textit{visual perception heads}. Increasing attention to visual features alone reduces model hallucinations, while confining this reallocation strategy to \textit{visual perception heads} minimizes adverse effects on content quality. More ablation studies can be found in \cref{app: ablation studies}.

\begin{table}[h]
\centering
\begin{tabular}{@{}c|c|cc@{}}
\toprule[1pt]
\toprule
\textbf{Model}                  & \textbf{Visual Restriction} & \textbf{POPE} & \textbf{Nocaps} \\ \midrule
\multirow{2}{*}{LLaVA-7B}  & \ding{51}                     & 89.8          & 78.8            \\
                                & \ding{55}                    & 89.9          & \textbf{76.4}   \\ \midrule
\multirow{2}{*}{LLaVA-13B} & \ding{51}                        & 90.2          & 82.3            \\
                                & \ding{55}                    & 90.0          & \textbf{81.1}   \\ \bottomrule
\bottomrule[1pt]
\end{tabular}
\caption{\textbf{Ablation Study of Visual Perception Restriction Mechanism.} Restricting attention redistribution to the \textit{visual perception heads} more effectively preserves the quality of generated content.}
\label{tab: visual restriction}
\end{table}
\vspace{-0.25cm}

%% file: sec/7_conclusion.tex
\section{conclusion}

In this paper, we identify two key drawbacks of using contrastive decoding to mitigate hallucinations in MLLMs: reduced quality of generated content and slower inference speed. To address these challenges, we propose a novel approach, Visual Amplification Fusion, which effectively mitigates hallucinations while preserving both inference speed and content generation quality. By enhancing the attention to visual features during modality fusion, VAF minimizes the over-reliance on language priors, ensuring a high degree of consistency between generated content and visual inputs. Extensive experiments across multiple benchmarks and MLLMs demonstrate that VAF provides a clear advantage in hallucination mitigation.

%% file: sec/X_suppl.tex
\clearpage
\setcounter{page}{1}
\maketitlesupplementary

\section{Additional Experimental Results}
\label{app: experimental results}

\vspace{-0.1cm}

\cref{app: results on mme} presents the additional experimental results across all tasks in the MME benchmark. \cref{app: results on pope} details the experimental outcomes on the three datasets within the POPE benchmark. \cref{app: inference speed} compares the inference speeds and memory usage of various methods on ScienceQA and Nocaps. \cref{app: case studies} highlights case studies of the VAF method on the LLaVA-Bench dataset.

\begin{figure*}[b]
  \centering
   \includegraphics[width=0.85\linewidth]{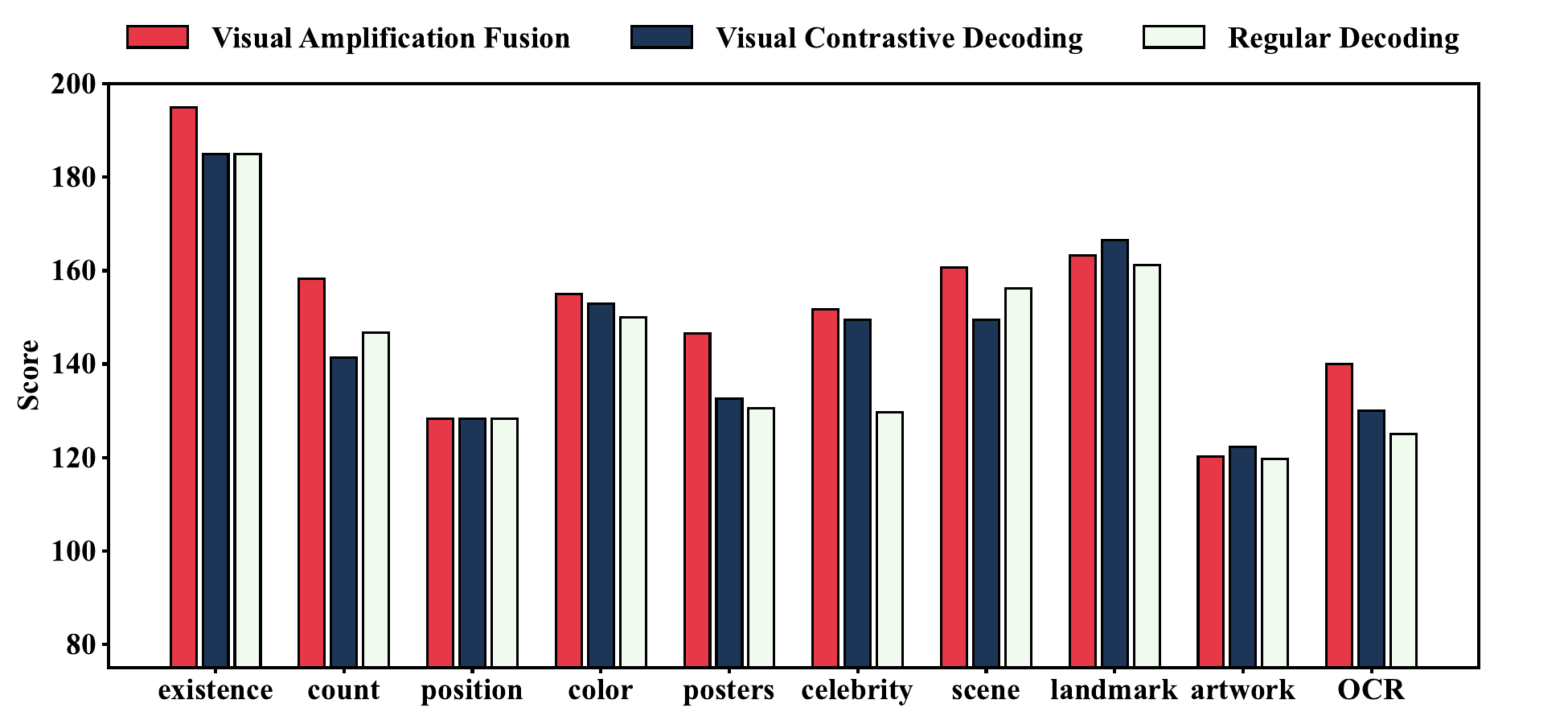}
   \vspace{-0.25cm}
   \caption{\textbf{Performance of LLaVA-v1.5-7B model on perception-related tasks in the MME Benchmark.} VAF consistently achieved the highest scores across nearly all perception tasks.}
   \label{fig: mme_p 7b}
\end{figure*}

\begin{figure*}[b]
  \centering
   \includegraphics[width=0.85\linewidth]{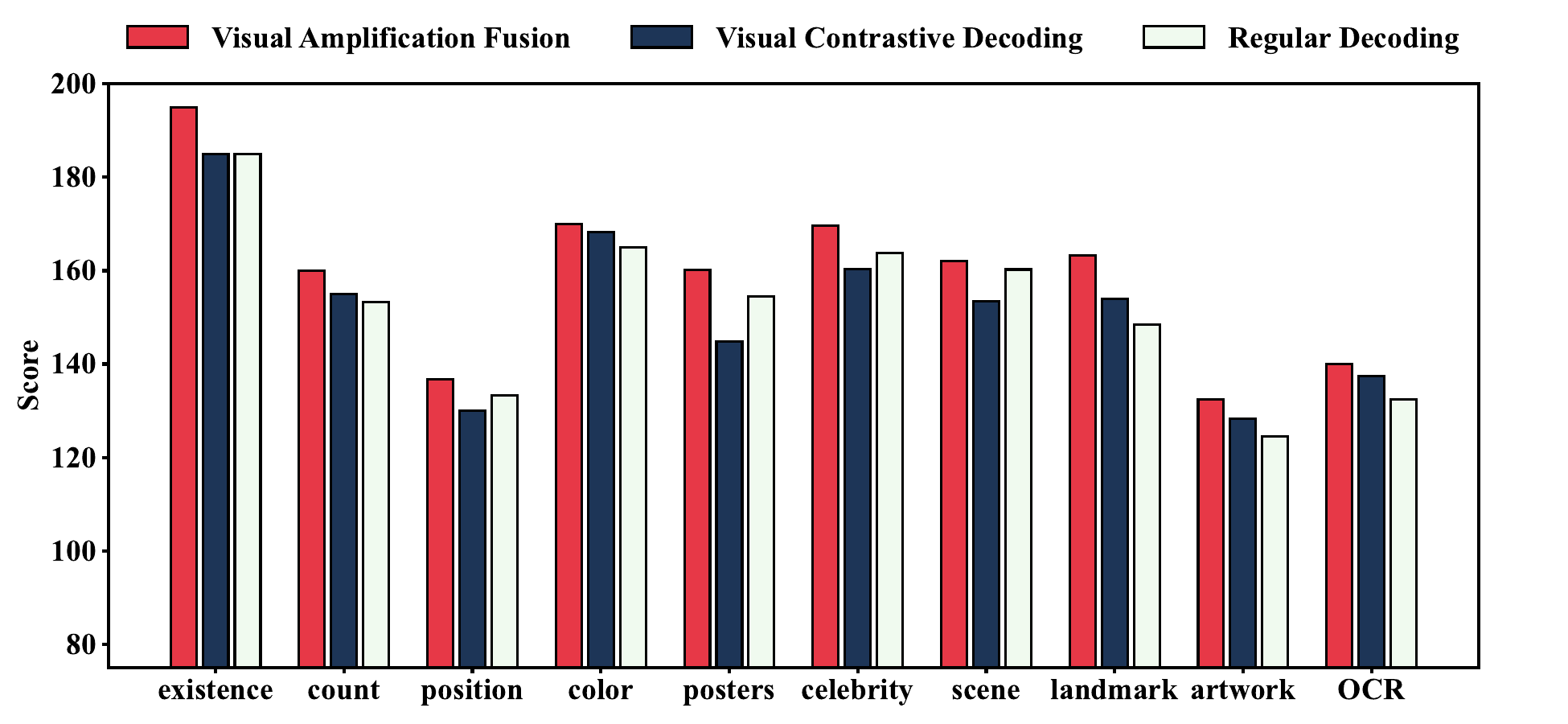}
   \vspace{-0.25cm}
   \caption{\textbf{Performance of LLaVA-v1.5-13B model on perception-related tasks in the MME Benchmark.} VAF consistently achieved the highest scores across nearly all perception tasks.}
   \label{fig: mme_p 13b}
\end{figure*}

\subsection{Detailed Experimental Results on MME}
\label{app: results on mme}

\cref{fig: mme_p 7b} and \cref{fig: mme_p 13b} present the performance of the LLaVA model family on perception-related tasks within the MME benchmark. Models utilizing the VAF method demonstrate significantly better performance compared to those employing the VCD method. Notably, VAF achieves consistent leadership across all tasks with the LLaVA-v1.5-13B model, likely due to its ability to balance attention between visual and language modalities, ensuring generated content aligns more closely with visual inputs.

\begin{figure*}[t]
  \centering
   \includegraphics[width=0.85\linewidth]{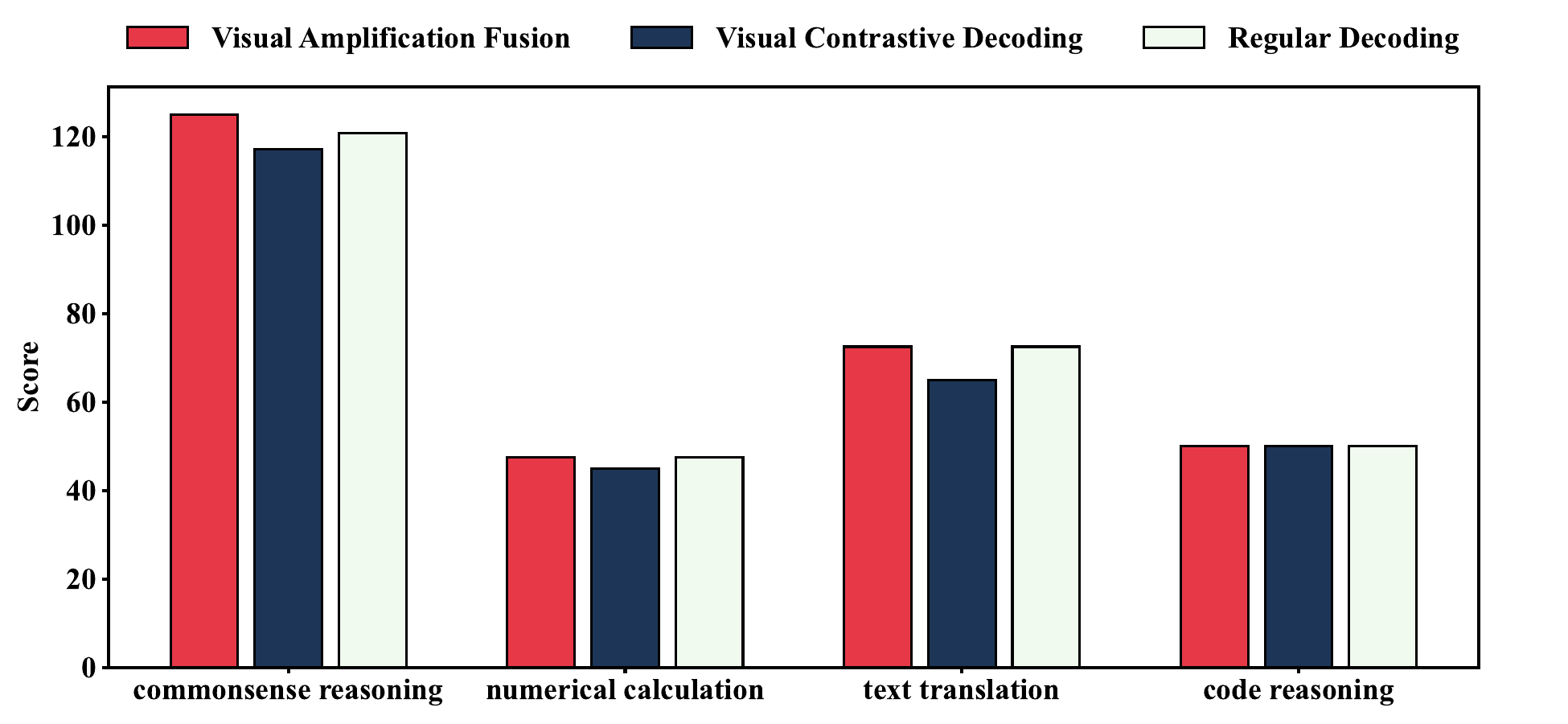}
   \vspace{-0.25cm}
   \caption{\textbf{Performance of the LLaVA-v1.5-7B model on cognition-related tasks in the MME Benchmark.} The VAF method delivers a slight performance improvement compared to the degradation observed with the VCD method.}
   \label{fig: mme_c 7b}
\end{figure*}

\begin{figure*}[t]
  \centering
   \includegraphics[width=0.85\linewidth]{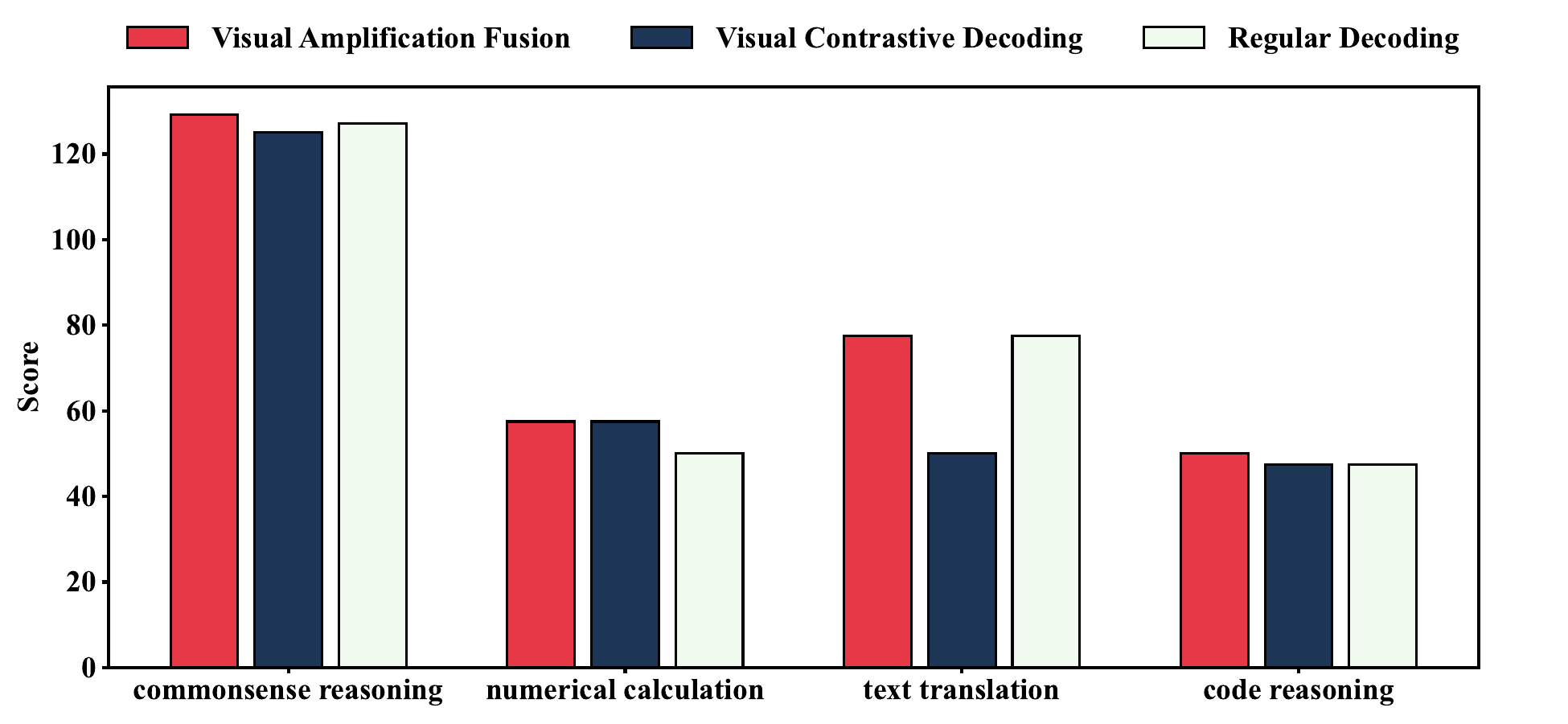}
   \vspace{-0.25cm}
   \caption{\textbf{Performance of the LLaVA-v1.5-13B model on cognition-related tasks in the MME Benchmark.} The VAF method delivers a slight performance improvement compared to the degradation observed with the VCD method.}
   \label{fig: mme_c 13b}
\end{figure*}

\cref{fig: mme_c 7b} and \cref{fig: mme_c 13b} illustrate the performance of LLaVA model family on cognition-related tasks within the MME benchmark. The application of the VCD method significantly impaired the model's performance on these tasks, likely due to its disruptive effect on linguistic priors. In contrast, VAF method not only avoided such negative impacts but also resulted in a slight performance improvement. This improvement is attributed to VAF’s ability to precisely resolve the model’s tendency to overlook visual features during the critical fusion stage, facilitating better integration of visual information while preserving its effective use of linguistic information.

\subsection{Detailed Experimental Results on POPE}
\label{app: results on pope}

\cref{tab: pope-llava-7b} and \cref{tab: pope-llava-13b} summarize the experimental results of the LLaVA-v.15 model family on the MSCOCO, A-OKVQA, and GQA datasets within the POPE benchmark. The results highlight that our approach consistently delivers more stable and significantly improved hallucination suppression compared to the VCD method. This advantage stems from our direct enhancement of attention to visual features during the modality fusion process, enabling balanced outputs across both visual and linguistic modalities. In contrast, the VCD method relies on suppressing language priors to indirectly enhance attention to visual information. Decoding method employed in all experiments utilizes greedy search.

\begin{table*}[t]
\centering
\begin{tabular}{@{}c|c|c|cccc@{}}
\toprule[1pt]
\toprule
\textbf{Dataset}          & \textbf{Category}             & \textbf{Method}             & \textbf{Accurancy}                                           & \textbf{Precision}                                           & \textbf{Recall}                                              & \textbf{F1-score}                                            \\ \midrule
                          &                               & Regular                     & 88.2                                                         & 94.2                                                         & 81.5                                                         & 87.4                                                         \\
                          &                               & VCD                         & 88.5                                                         & 94.4                                                         & 81.8                                                         & 87.6                                                         \\
                          & \multirow{-3}{*}{Random}      & \cellcolor[HTML]{EFEFEF}VAF & \cellcolor[HTML]{EFEFEF}{\color[HTML]{CB0000} \textbf{89.8}} & \cellcolor[HTML]{EFEFEF}92.9                                 & \cellcolor[HTML]{EFEFEF}86.2                                 & \cellcolor[HTML]{EFEFEF}{\color[HTML]{CB0000} \textbf{89.4}} \\ \cmidrule(l){2-7} 
                          &                               & Regular                     & 86.1                                                         & 89.9                                                         & 81.5                                                         & 85.5                                                         \\
                          &                               & VCD                         & 86.3                                                         & 90.0                                                         & 81.7                                                         & 85.8                                                         \\
                          & \multirow{-3}{*}{Popular}     & \cellcolor[HTML]{EFEFEF}VAF & \cellcolor[HTML]{EFEFEF}{\color[HTML]{CB0000} \textbf{87.5}} & \cellcolor[HTML]{EFEFEF}88.6                                 & \cellcolor[HTML]{EFEFEF}86.2                                 & \cellcolor[HTML]{EFEFEF}{\color[HTML]{CB0000} \textbf{87.4}} \\ \cmidrule(l){2-7} 
                          &                               & Regular                     & 82.3                                                         & 82.9                                                         & 81.3                                                         & 82.1                                                         \\
                          &                               & VCD                         & 82.3                                                         & 82.9                                                         & 81.6                                                         & 82.4                                                         \\
\multirow{-9}{*}{MSCOCO}  & \multirow{-3}{*}{Adverserial} & \cellcolor[HTML]{EFEFEF}VAF & \cellcolor[HTML]{EFEFEF}{\color[HTML]{CB0000} \textbf{83.4}} & \cellcolor[HTML]{EFEFEF}86.8                                 & \cellcolor[HTML]{EFEFEF}78.9                                 & \cellcolor[HTML]{EFEFEF}{\color[HTML]{CB0000} \textbf{82.6}} \\ \midrule
                          &                               & Regular                     & 87.6                                                         & 87.6                                                         & 87.7                                                         & 87.6                                                         \\
                          &                               & VCD                         & 87.7                                                         & 87.8                                                         & 87.6                                                         & 87.8                                                         \\
                          & \multirow{-3}{*}{Random}      & \cellcolor[HTML]{EFEFEF}VAF & \cellcolor[HTML]{EFEFEF}{\color[HTML]{CB0000} \textbf{89.4}} & \cellcolor[HTML]{EFEFEF}91.7                                 & \cellcolor[HTML]{EFEFEF}86.6                                 & \cellcolor[HTML]{EFEFEF}{\color[HTML]{CB0000} \textbf{89.1}} \\ \cmidrule(l){2-7} 
                          &                               & Regular                     & 81.9                                                         & 78.4                                                         & 87.7                                                         & 82.8                                                         \\
                          &                               & VCD                         & 82.1                                                         & 78.5                                                         & 87.9                                                         & 83.1                                                         \\
                          & \multirow{-3}{*}{Popular}     & \cellcolor[HTML]{EFEFEF}VAF & \cellcolor[HTML]{EFEFEF}{\color[HTML]{CB0000} \textbf{84.2}} & \cellcolor[HTML]{EFEFEF}82.6                                 & \cellcolor[HTML]{EFEFEF}86.6                                 & \cellcolor[HTML]{EFEFEF}{\color[HTML]{CB0000} \textbf{84.6}} \\ \cmidrule(l){2-7} 
                          &                               & Regular                     & 74.3                                                         & 68.8                                                         & 87.7                                                         & 77.1                                                         \\
                          &                               & VCD                         & 72.4                                                         & 68.0                                                         & 87.4                                                         & 76.7                                                         \\
\multirow{-9}{*}{A-OKVQA} & \multirow{-3}{*}{Adverserial} & \cellcolor[HTML]{EFEFEF}VAF & \cellcolor[HTML]{EFEFEF}{\color[HTML]{CB0000} \textbf{77.2}} & \cellcolor[HTML]{EFEFEF}72.9                                 & \cellcolor[HTML]{EFEFEF}86.6                                 & \cellcolor[HTML]{EFEFEF}{\color[HTML]{CB0000} \textbf{79.2}} \\ \midrule
                          &                               & Regular                     & 88.0                                                         & 87.1                                                         & 89.3                                                         & 88.2                                                         \\
                          &                               & VCD                         & 88.6                                                         & 87.4                                                         & 89.5                                                         & 88.8                                                         \\
                          & \multirow{-3}{*}{Random}      & \cellcolor[HTML]{EFEFEF}VAF & \cellcolor[HTML]{EFEFEF}{\color[HTML]{CB0000} \textbf{89.5}} & \cellcolor[HTML]{EFEFEF}90.8                                 & \cellcolor[HTML]{EFEFEF}88.0                                 & \cellcolor[HTML]{EFEFEF}{\color[HTML]{CB0000} \textbf{89.4}} \\ \cmidrule(l){2-7} 
                          &                               & Regular                     & 79.4                                                         & 74.4                                                         & 89.3                                                         & 81.1                                                         \\
                          &                               & VCD                         & 79.9                                                         & 74.6                                                         & 89.5                                                         & 81.7                                                         \\
                          & \multirow{-3}{*}{Popular}     & \cellcolor[HTML]{EFEFEF}VAF & \cellcolor[HTML]{EFEFEF}{\color[HTML]{CB0000} \textbf{81.8}} & \cellcolor[HTML]{EFEFEF}78.3                                 & \cellcolor[HTML]{EFEFEF}88.0                                 & \cellcolor[HTML]{EFEFEF}{\color[HTML]{CB0000} \textbf{82.9}} \\ \cmidrule(l){2-7} 
                          &                               & Regular                     & 76.3                                                         & 70.6                                                         & 89.3                                                         & 78.9                                                         \\
                          &                               & VCD                         & 75.2                                                         & 70.2                                                         & 89.9                                                         & 78.3                                                         \\
\multirow{-9}{*}{GQA}     & \multirow{-3}{*}{Adverserial} & \cellcolor[HTML]{EFEFEF}VAF & \cellcolor[HTML]{EFEFEF}{\color[HTML]{CB0000} \textbf{79.7}} & \cellcolor[HTML]{EFEFEF}75.4                                 & \cellcolor[HTML]{EFEFEF}88.0                                 & \cellcolor[HTML]{EFEFEF}{\color[HTML]{CB0000} \textbf{81.2}} \\ \bottomrule
\bottomrule[1pt]
\end{tabular}
\caption{\textbf{Experimental results of LLaVA-1.5-7B model on POPE.} VAF method achieves the most effective hallucination suppression across all three datasets. For emphasis, the {\color[HTML]{CB0000} \textbf{highest scores}} in each setting are highlighted in {\color[HTML]{CB0000} \textbf{red}}.}
\label{tab: pope-llava-7b}
\end{table*}

\begin{table*}[t]
\centering
\begin{tabular}{@{}c|c|cccc@{}}
\toprule[1pt]
\toprule
\textbf{Model}                   & \textbf{Method}             & \textbf{Accurancy}           & \textbf{Total Time}                   & \textbf{GPU-Memory}                   & \textbf{Latency/Example}               \\ \midrule
                                 & Regular                     & 88.2                         & 5:32                                  & 14.5G                                 & 0.111s                                 \\
                                 & VCD                         & 88.5                         & {\color[HTML]{CB0000} \textbf{10:31}} & {\color[HTML]{CB0000} \textbf{15.7G}} & {\color[HTML]{CB0000} \textbf{0.210s}} \\
\multirow{-3}{*}{LLaVA-v1.5-7B}  & \cellcolor[HTML]{EFEFEF}VAF & \cellcolor[HTML]{EFEFEF}89.8 & \cellcolor[HTML]{EFEFEF}5:48          & \cellcolor[HTML]{EFEFEF}14.5G         & \cellcolor[HTML]{EFEFEF}0.116s         \\ \midrule
                                 & Regular                     & 88.4                         & 8:39                                  & 26.7G                                 & 0.173s                                 \\
                                 & VCD                         & 88.6                         & {\color[HTML]{CB0000} \textbf{19:38}} & {\color[HTML]{CB0000} \textbf{27.8G}} & {\color[HTML]{CB0000} \textbf{0.392s}} \\
\multirow{-3}{*}{LLaVA-v1.5-13B} & \cellcolor[HTML]{EFEFEF}VAF & \cellcolor[HTML]{EFEFEF}90.2 & \cellcolor[HTML]{EFEFEF}8:45          & \cellcolor[HTML]{EFEFEF}26.7G         & \cellcolor[HTML]{EFEFEF}0.175s         \\ \bottomrule
\bottomrule[1pt]
\end{tabular}
\caption{\textbf{A comparison of inference speed and GPU memory usage for different methods applied to the LLaVA-v1.5 model family on POPE benchmark.} Results with the slowest inference speed and highest memory usage are highlighted in {\color[HTML]{CB0000} \textbf{red}}.}
\label{tab: inference pope}
\end{table*}

\begin{table*}[b]
\centering
\begin{tabular}{@{}c|c|cccc@{}}
\toprule[1pt]
\toprule
\textbf{Model}                   & \textbf{Method}             & \textbf{Accurancy}           & \textbf{Total Time}                     & \textbf{GPU-Memory}                   & \textbf{Latency/Example}               \\ \midrule
                                 & Regular                     & 68.0                         & 0:36:39                                 & 14.5G                                 & 0.488s                                 \\
                                 & VCD                         & 64.5                         & {\color[HTML]{CB0000} \textbf{1:18:47}} & {\color[HTML]{CB0000} \textbf{15.7G}} & {\color[HTML]{CB0000} \textbf{1.058s}} \\
\multirow{-3}{*}{LLaVA-v1.5-7B}  & \cellcolor[HTML]{EFEFEF}VAF & \cellcolor[HTML]{EFEFEF}68.5 & \cellcolor[HTML]{EFEFEF}0:36:41         & \cellcolor[HTML]{EFEFEF}14.5G         & \cellcolor[HTML]{EFEFEF}0.489s          \\ \midrule
                                 & Regular                     & 71.6                         & 0:45:20                                 & 26.7G                                 & 0.604s                                 \\
                                 & VCD                         & 70.0                         & {\color[HTML]{CB0000} \textbf{1:46:59}} & {\color[HTML]{CB0000} \textbf{27.8G}} & {\color[HTML]{CB0000} \textbf{1.426s}} \\
\multirow{-3}{*}{LLaVA-v1.5-13B} & \cellcolor[HTML]{EFEFEF}VAF & \cellcolor[HTML]{EFEFEF}71.7 & \cellcolor[HTML]{EFEFEF}0:48:24         & \cellcolor[HTML]{EFEFEF}26.7G         & \cellcolor[HTML]{EFEFEF}0.645s         \\ \bottomrule
\bottomrule[1pt]
\end{tabular}
\caption{\textbf{A comparison of inference speed and GPU memory usage for different methods applied to the LLaVA-v1.5 model family on Nocaps benchmark.} Results with the slowest inference speed and highest memory usage are highlighted in {\color[HTML]{CB0000} \textbf{red}}.}
\label{tab: inference nocaps}
\end{table*}

\begin{table*}[b]
\centering
\begin{tabular}{@{}c|c|c|cccc@{}}
\toprule[1pt]
\toprule
\textbf{Dataset}          & \textbf{Category}             & \textbf{Method}                                    & \textbf{Accurancy}                                           & \textbf{Precision}                                           & \textbf{Recall}                                              & \textbf{F1-score}                                            \\ \midrule
                          &                               & Regular                                            & 88.4                                                         & 94.6                                                         & 81.6                                                         & 87.6                                                         \\
                          &                               & VCD                                                & 88.6                                                         & 95.0                                                         & 81.8                                                         & 87.7                                                         \\
                          & \multirow{-3}{*}{Random}      & \cellcolor[HTML]{EFEFEF}{\color[HTML]{000000} VAF} & \cellcolor[HTML]{EFEFEF}{\color[HTML]{CB0000} \textbf{90.2}} & \cellcolor[HTML]{EFEFEF}94.2                                 & \cellcolor[HTML]{EFEFEF}85.6                                 & \cellcolor[HTML]{EFEFEF}{\color[HTML]{CB0000} \textbf{89.7}} \\ \cmidrule(l){2-7} 
                          &                               & Regular                                            & 86.9                                                         & 91.3                                                         & 81.6                                                         & 86.2                                                         \\
                          &                               & VCD                                                & 87.0                                                         & 91.4                                                         & 82.0                                                         & 86.4                                                         \\
                          & \multirow{-3}{*}{Popular}     & \cellcolor[HTML]{EFEFEF}{\color[HTML]{000000} VAF} & \cellcolor[HTML]{EFEFEF}{\color[HTML]{CB0000} \textbf{88.4}} & \cellcolor[HTML]{EFEFEF}90.6                                 & \cellcolor[HTML]{EFEFEF}85.6                                 & \cellcolor[HTML]{EFEFEF}{\color[HTML]{CB0000} \textbf{88.0}} \\ \cmidrule(l){2-7} 
                          &                               & Regular                                            & 83.4                                                         & 84.9                                                         & 81.4                                                         & 83.1                                                         \\
                          &                               & VCD                                                & 83.7                                                         & 85.1                                                         & 81.7                                                         & 83.1                                                         \\
\multirow{-9}{*}{MSCOCO}  & \multirow{-3}{*}{Adverserial} & \cellcolor[HTML]{EFEFEF}{\color[HTML]{000000} VAF} & \cellcolor[HTML]{EFEFEF}{\color[HTML]{CB0000} \textbf{84.5}} & \cellcolor[HTML]{EFEFEF}83.8                                 & \cellcolor[HTML]{EFEFEF}85.5                                 & \cellcolor[HTML]{EFEFEF}{\color[HTML]{CB0000} \textbf{84.7}} \\ \midrule
                          &                               & Regular                                            & 88.0                                                         & 88.8                                                         & 87.1                                                         & 87.9                                                         \\
                          &                               & VCD                                                & 88.2                                                         & 89.2                                                         & 87.5                                                         & 87.9                                                         \\
                          & \multirow{-3}{*}{Random}      & \cellcolor[HTML]{EFEFEF}{\color[HTML]{000000} VAF} & \cellcolor[HTML]{EFEFEF}{\color[HTML]{CB0000} \textbf{89.4}} & \cellcolor[HTML]{EFEFEF}91.4                                 & \cellcolor[HTML]{EFEFEF}86.8                                 & \cellcolor[HTML]{EFEFEF}{\color[HTML]{CB0000} \textbf{89.1}} \\ \cmidrule(l){2-7} 
                          &                               & Regular                                            & 83.9                                                         & 81.7                                                         & 87.1                                                         & 84.3                                                         \\
                          &                               & VCD                                                & 84.2                                                         & 81.7                                                         & 87.3                                                         & 84.3                                                         \\
                          & \multirow{-3}{*}{Popular}     & \cellcolor[HTML]{EFEFEF}{\color[HTML]{000000} VAF} & \cellcolor[HTML]{EFEFEF}{\color[HTML]{CB0000} \textbf{86.0}} & \cellcolor[HTML]{EFEFEF}85.4                                 & \cellcolor[HTML]{EFEFEF}86.8                                 & \cellcolor[HTML]{EFEFEF}{\color[HTML]{CB0000} \textbf{86.1}} \\ \cmidrule(l){2-7} 
                          &                               & Regular                                            & 76.0                                                         & 71.0                                                         & 87.1                                                         & 78.2                                                         \\
                          &                               & VCD                                                & 76.4                                                         & 71.2                                                         & 87.1                                                         & 78.3                                                         \\
\multirow{-9}{*}{A-OKVQA} & \multirow{-3}{*}{Adverserial} & \cellcolor[HTML]{EFEFEF}{\color[HTML]{000000} VAF} & \cellcolor[HTML]{EFEFEF}{\color[HTML]{CB0000} \textbf{78.2}} & \cellcolor[HTML]{EFEFEF}74.1                                 & \cellcolor[HTML]{EFEFEF}86.8                                 & \cellcolor[HTML]{EFEFEF}{\color[HTML]{CB0000} \textbf{79.9}} \\ \midrule
                          &                               & Regular                                            & 88.3                                                         & 87.8                                                         & 89.0                                                         & 88.4                                                         \\
                          &                               & VCD                                                & 88.3                                                         & 88.1                                                         & 89.3                                                         & 88.5                                                         \\
                          & \multirow{-3}{*}{Random}      & \cellcolor[HTML]{EFEFEF}{\color[HTML]{000000} VAF} & \cellcolor[HTML]{EFEFEF}{\color[HTML]{CB0000} \textbf{89.7}} & \cellcolor[HTML]{EFEFEF}87.8                                 & \cellcolor[HTML]{EFEFEF}92.2                                 & \cellcolor[HTML]{EFEFEF}{\color[HTML]{CB0000} \textbf{89.9}} \\ \cmidrule(l){2-7} 
                          &                               & Regular                                            & 83.3                                                         & 79.8                                                         & 89.0                                                         & 84.1                                                         \\
                          &                               & VCD                                                & 83.2                                                         & 80.0                                                         & 89.2                                                         & 84.1                                                         \\
                          & \multirow{-3}{*}{Popular}     & \cellcolor[HTML]{EFEFEF}{\color[HTML]{000000} VAF} & \cellcolor[HTML]{EFEFEF}{\color[HTML]{CB0000} \textbf{85.2}} & \cellcolor[HTML]{EFEFEF}83.0                                 & \cellcolor[HTML]{EFEFEF}88.6                                 & \cellcolor[HTML]{EFEFEF}{\color[HTML]{CB0000} \textbf{85.7}} \\ \cmidrule(l){2-7} 
                          &                               & Regular                                            & 78.5                                                         & 73.3                                                         & 89.0                                                         & 80.4                                                         \\
                          &                               & VCD                                                & 78.7                                                         & 73.3                                                         & 88.9                                                         & 80.3                                                         \\
\multirow{-9}{*}{GQA}     & \multirow{-3}{*}{Adverserial} & \cellcolor[HTML]{EFEFEF}{\color[HTML]{000000} VAF} & \cellcolor[HTML]{EFEFEF}{\color[HTML]{CB0000} \textbf{80.8}} & \cellcolor[HTML]{EFEFEF}76.6                                 & \cellcolor[HTML]{EFEFEF}88.6                                 & \cellcolor[HTML]{EFEFEF}{\color[HTML]{CB0000} \textbf{82.1}} \\ \bottomrule
\bottomrule[1pt]
\end{tabular}
\caption{\textbf{Experimental results of LLaVA-1.5-13B model on POPE.} VAF method achieves the most effective hallucination suppression across all three datasets. For emphasis, the {\color[HTML]{CB0000} \textbf{highest scores}} in each setting are highlighted in {\color[HTML]{CB0000} \textbf{red}}.}
\label{tab: pope-llava-13b}
\end{table*}

\subsection{Comparison of Inference Speeds}
\label{app: inference speed}

\cref{tab: inference pope} and \cref{tab: inference nocaps} assess the impact of various methods on the LLaVA-v1.5 model family, focusing on inference speed and GPU memory usage. The results indicate that VCD significantly slows down inference, whereas our proposed method has a minimal effect. Furthermore, our method introduces no additional GPU memory requirements, in contrast to VCD, which incurs substantial GPU memory overhead. This efficiency is achieved because our approach eliminates the need for extra processing of contrastive inputs, thereby significantly reducing computational overhead. All experiments were performed on a server equipped with a single A800 80G GPU, employing greedy search as the decoding strategy.

\begin{figure*}[t]
  \centering
   \includegraphics[width=0.8\linewidth]{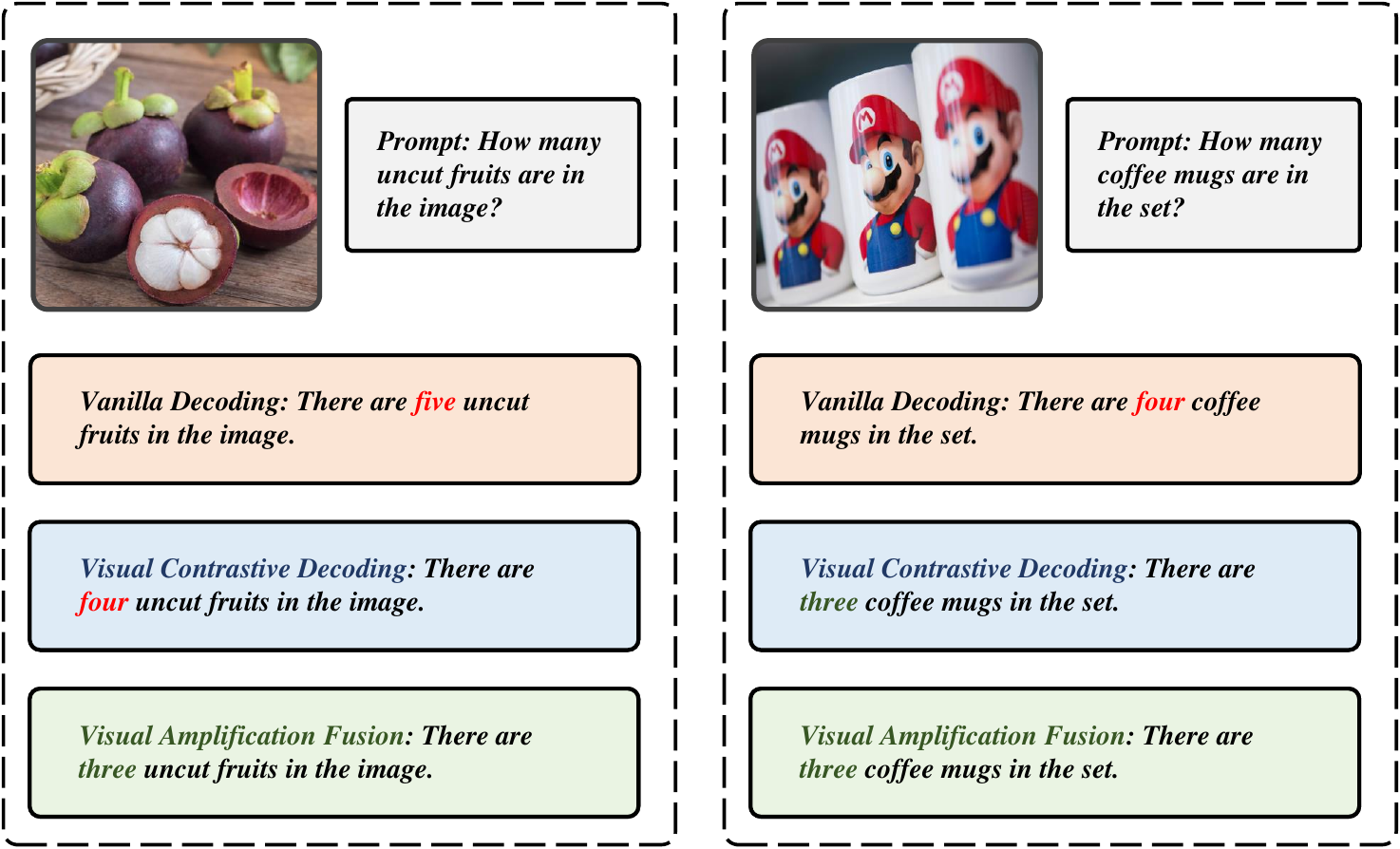}
   \caption{\textbf{An illustration of VAF correcting hallucinations on LLaVA-Bench, with a focus on numerical perception tasks.} {\color[HTML]{036400} \textbf{Correct}} outputs are highlighted in {\color[HTML]{036400} \textbf{green}}, while {\color[HTML]{CB0000} \textbf{incorrect}} ones are marked in {\color[HTML]{CB0000} \textbf{red}}.}
   \label{fig: llava count}
\end{figure*}

\begin{figure*}[t]
  \centering
   \includegraphics[width=0.8\linewidth]{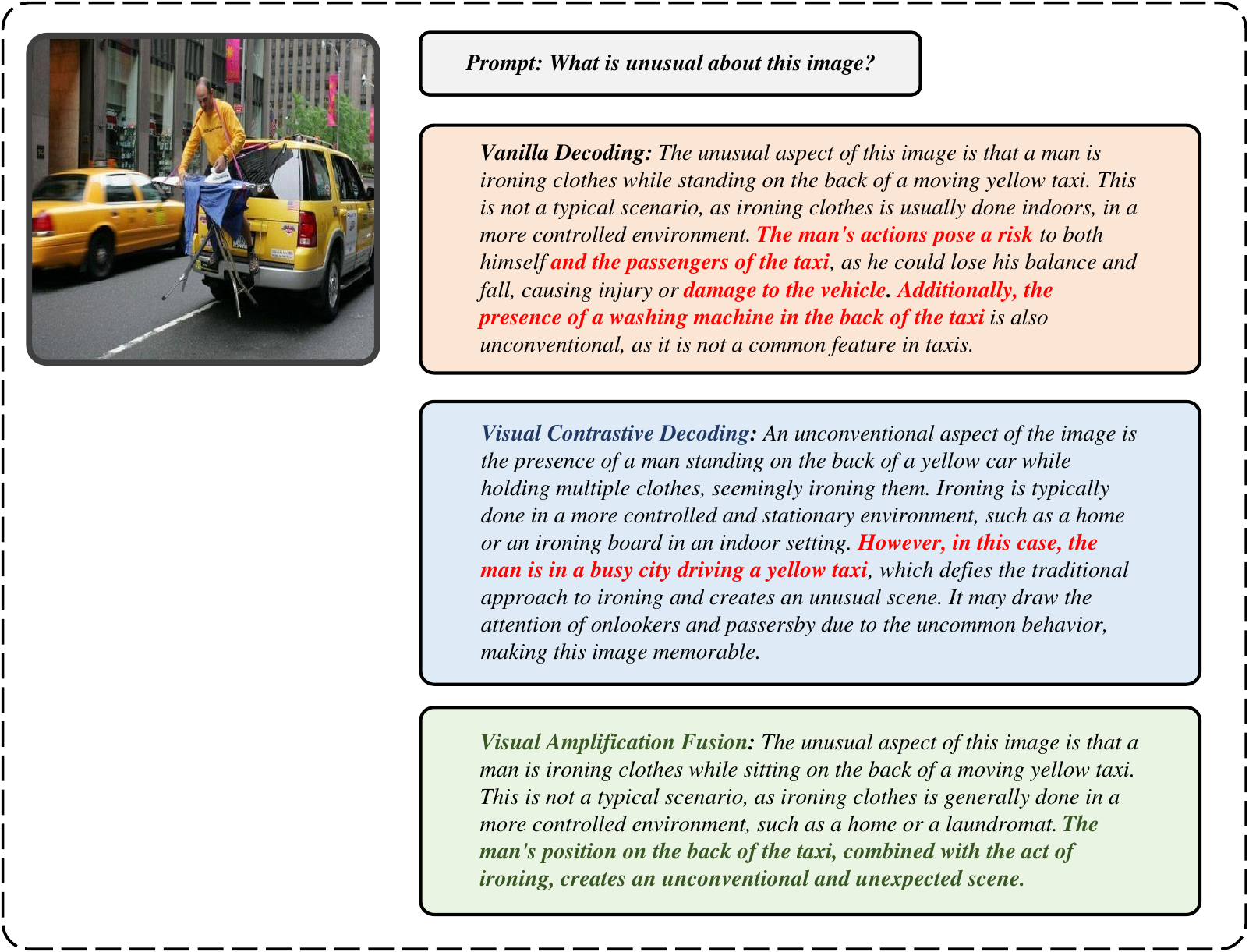}
   \caption{\textbf{An illustration of VAF correcting hallucinations on LLaVA-Bench, with a focus on complex reasoning tasks.} {\color[HTML]{036400} \textbf{Correct}} outputs are highlighted in {\color[HTML]{036400} \textbf{green}}, while {\color[HTML]{CB0000} \textbf{incorrect}} ones are marked in {\color[HTML]{CB0000} \textbf{red}}.}
   \label{fig: llava reason}
\end{figure*}

\begin{figure*}[t]
  \centering
   \includegraphics[width=0.71\linewidth]{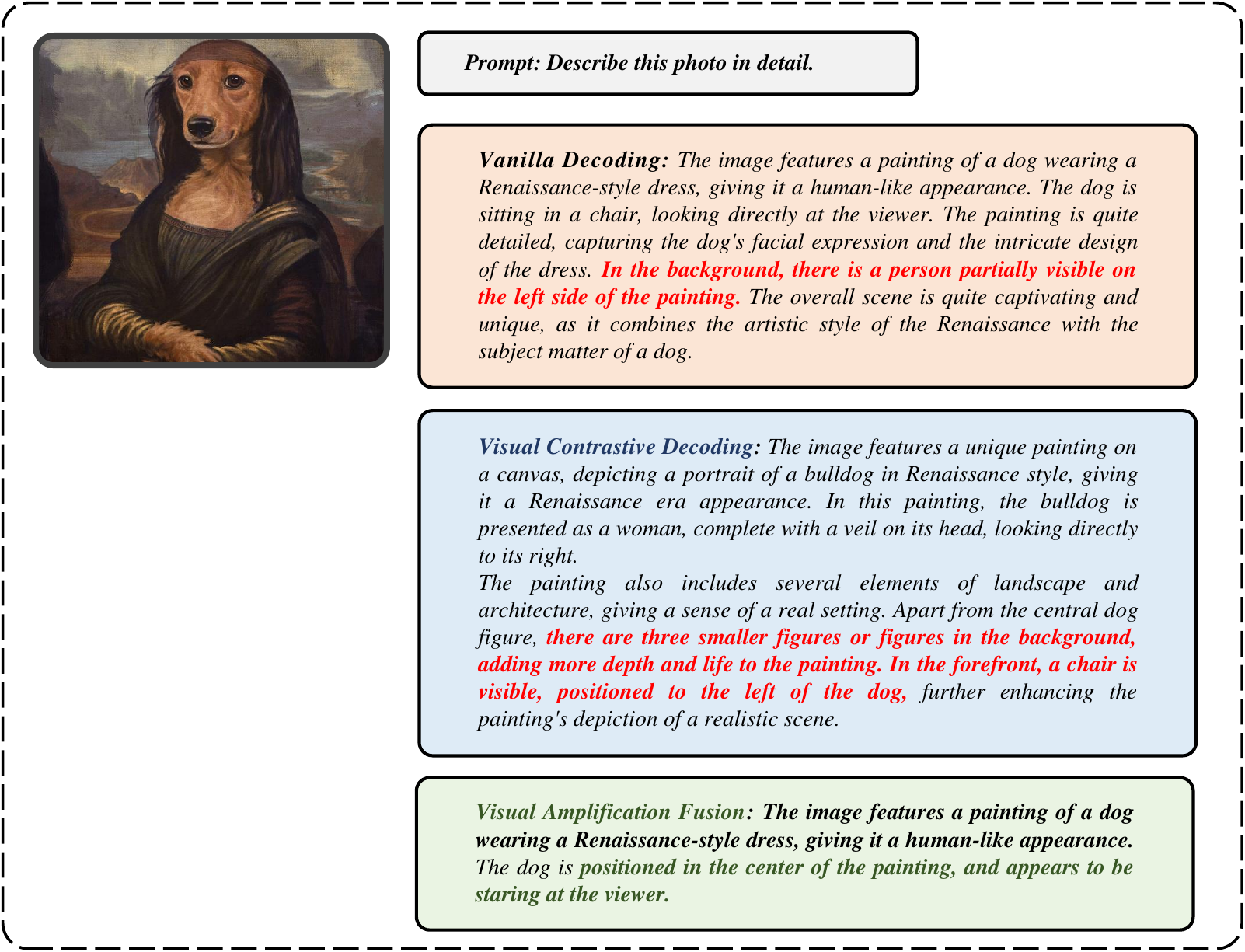}
   \caption{\textbf{An illustration of VAF correcting hallucinations on LLaVA-Bench, with a focus on image description tasks.} {\color[HTML]{036400} \textbf{Correct}} outputs are highlighted in {\color[HTML]{036400} \textbf{green}}, while {\color[HTML]{CB0000} \textbf{incorrect}} ones are marked in {\color[HTML]{CB0000} \textbf{red}}.}
   \label{fig: llava describe1}
\end{figure*}

\begin{figure*}[t]
  \centering
   \includegraphics[width=0.71\linewidth]{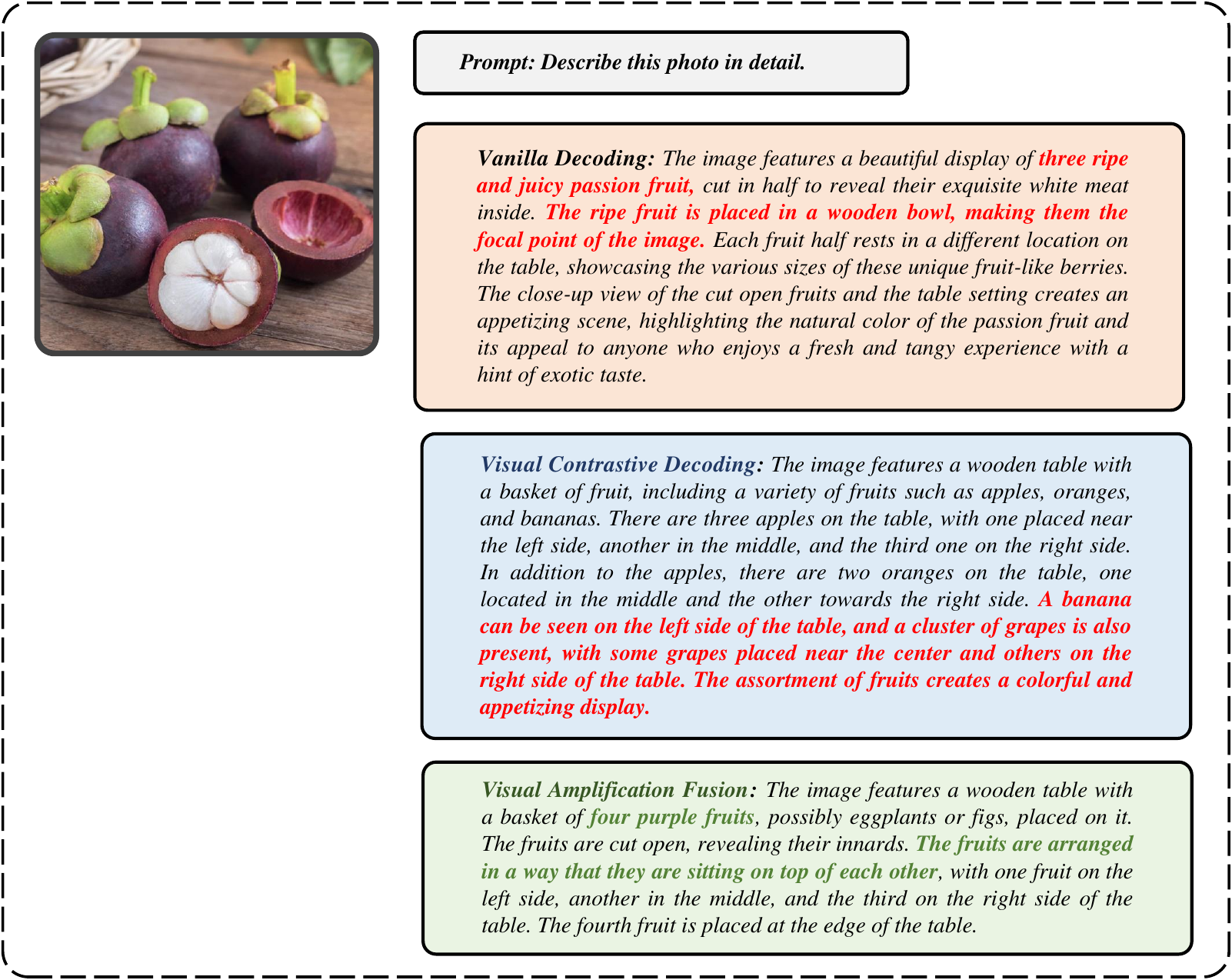}
   \caption{Additional example of VAF correcting model hallucinations on LLaVA-Bench, primarily focusing on image description tasks. {\color[HTML]{036400} \textbf{Correct}} outputs are highlighted in {\color[HTML]{036400} \textbf{green}}, while {\color[HTML]{CB0000} \textbf{incorrect}} ones are marked in {\color[HTML]{CB0000} \textbf{red}}.}
   \label{fig: llava describe2}
\end{figure*}

\begin{figure*}[ht]
    \centering
    \begin{subfigure}[b]{0.47\textwidth}
        \centering
        \includegraphics[width=\textwidth]{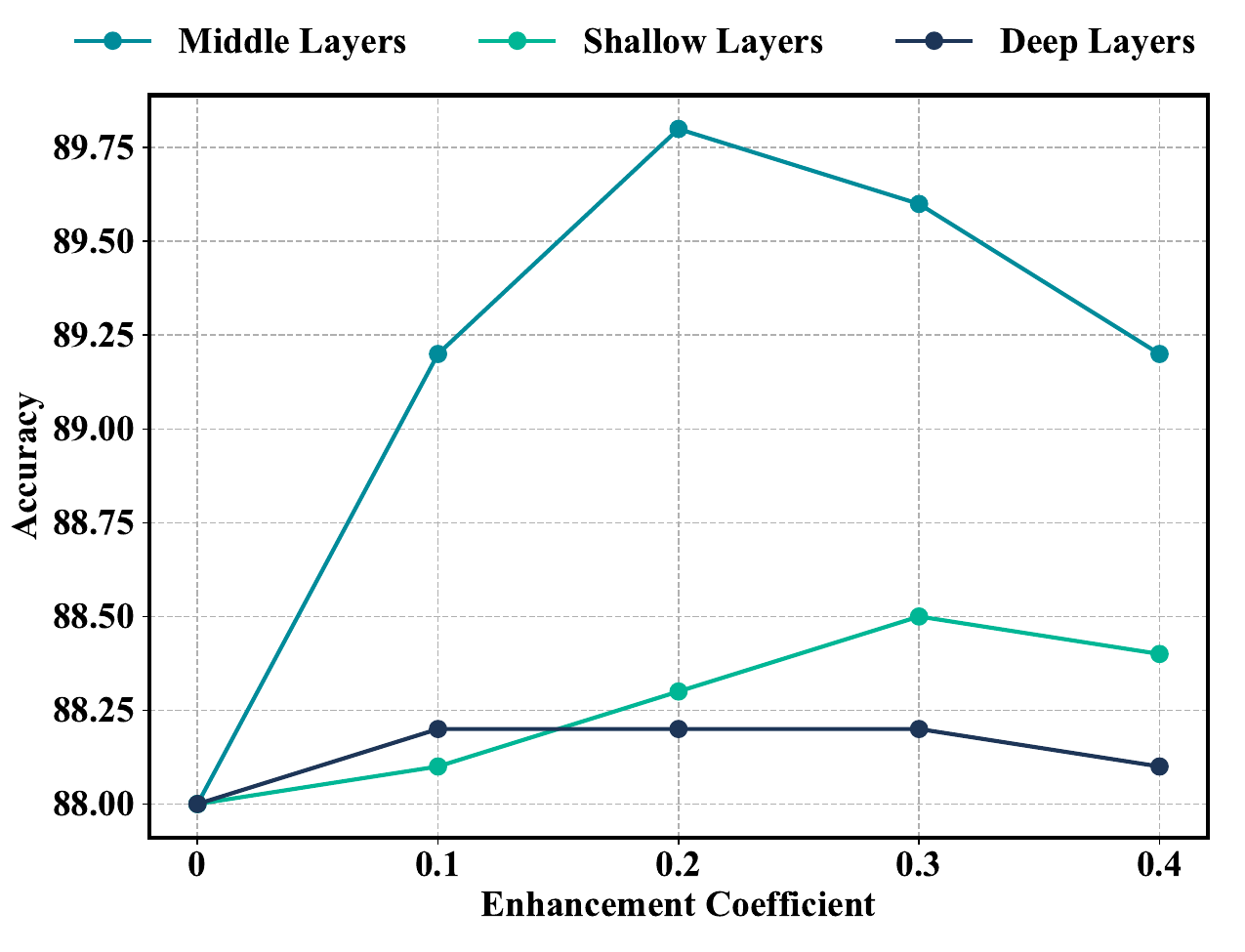}
        \caption{Accuracy Metric}
        \label{fig: layer acc}
    \end{subfigure}
    \hspace{0.05\textwidth}
    \begin{subfigure}[b]{0.47\textwidth}
        \centering
        \includegraphics[width=\textwidth]{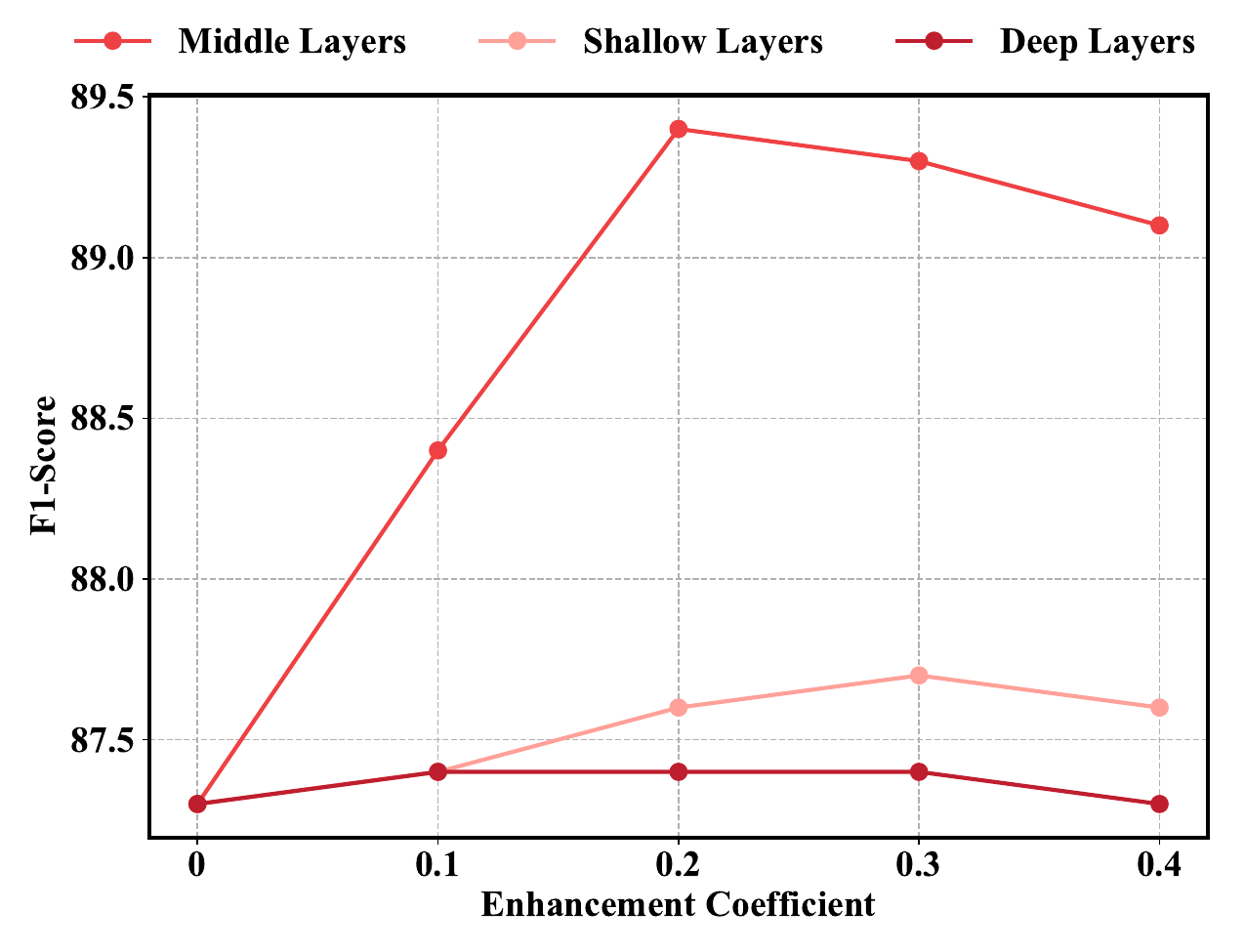}
        \caption{F1-Score Metirc}
        \label{fig: layer f1}
    \end{subfigure}
    \caption{\textbf{The Effect of Enhancing Visual Attention at Different Layers on Prediction Accuracy.} This experiment, conducted with the LLaVA-v1.5-7B model on the COCO-Random dataset within the POPE Benchmark, demonstrates that enhancing attention to visual features in the model's middle layers significantly reduces hallucinations.}
    \label{fig: enhance layer}
\end{figure*}

\subsection{Case study on LLaVA-Bench}
\label{app: case studies}

\cref{fig: llava count}, \cref{fig: llava reason}, \cref{fig: llava describe1}, and \cref{fig: llava describe2} illustrate the effectiveness of various methods in mitigating model hallucinations on LLaVA-Bench. Across tasks such as numerical perception, image description, and complex reasoning, our approach demonstrates consistently superior performance in suppressing hallucinations. Experiments are conducted using LLaVA-v1.5-7B model.

\section{Additional Ablation Studies}
\label{app: ablation studies}

In \cref{app: enhance layer}, we examine how enhancing attention to visual features at different levels affects hallucination suppression. In \cref{app: beta value}, we analyze the influence of varying the suppression coefficient $\beta$ on mitigating hallucinations. Finally, in \cref{app: various sampling}, we evaluate the performance of the VAF method in suppressing hallucinations under various sampling strategies.

\subsection{Effect of Enhancement at Different Layers}
\label{app: enhance layer}

We enhanced attention to visual features in layers 0-5, 10-15, and 20-25. \cref{fig: enhance layer} demonstrates the impact of enhancing visual attention at different layers. Notably, enhancing attention in the middle layers significantly reduces hallucination, while modifications in the shallow and deep layers have minimal effect on the generation results. As discussed in \cref{subsec: mid-layer}, this is because the model primarily integrates modality information in the middle layers. Thus, enhancing the focus on visual features during this phase is crucial for effectively mitigating hallucination. Experiments are conducted using LLaVA-v1.5-7B model on COCO-Random dataset from the POPE Benchmark.

\begin{figure}[ht]
  \centering
   \includegraphics[width=0.9\linewidth]{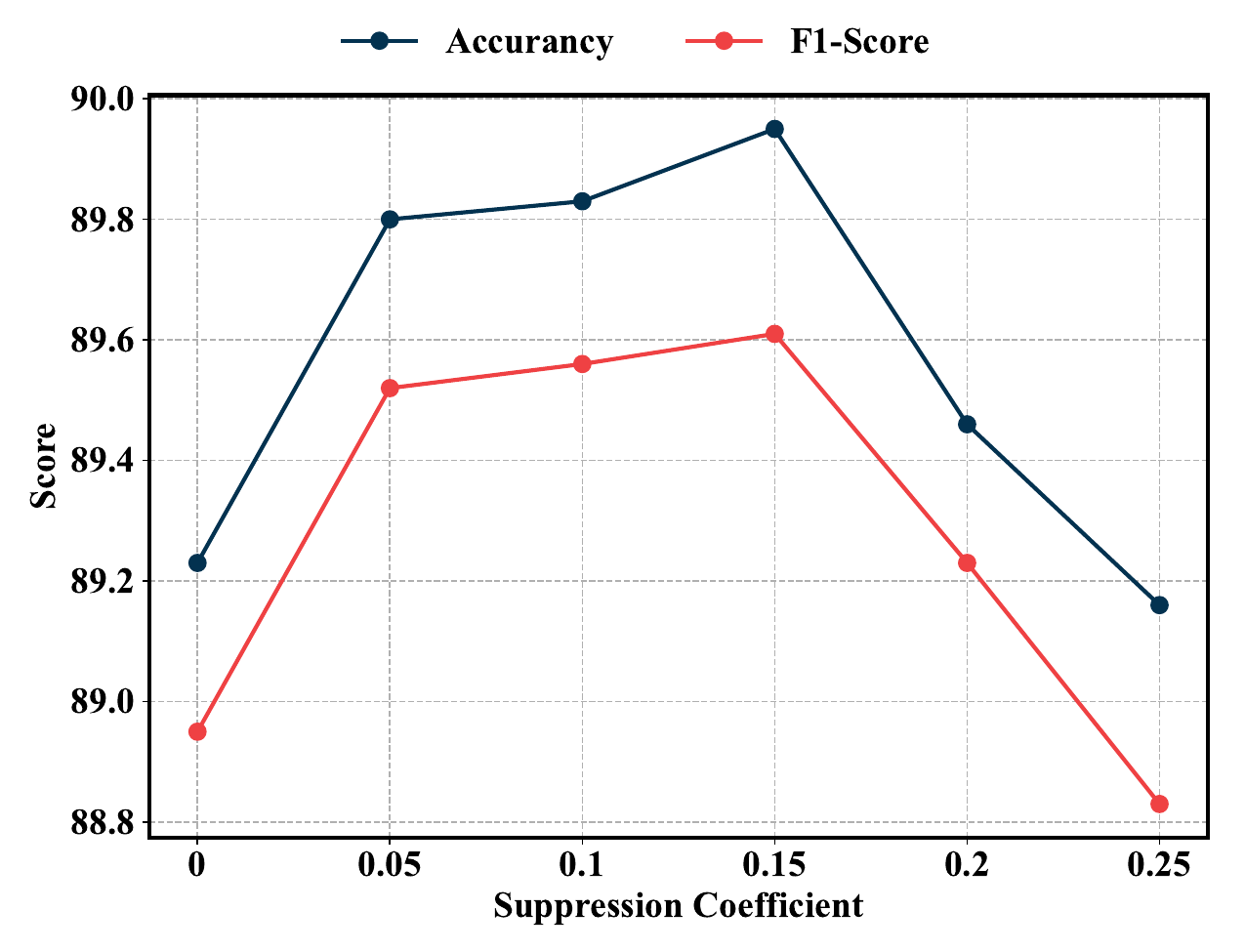}
   \caption{\textbf{The effect of the suppression coefficient $\beta$ on the VAF method's ability to mitigate model hallucinations.} The experiments were performed using the LLaVA-v1.5-7B model on the COCO-Random dataset from the POPE Benchmark.}
   \label{fig: beta}
\end{figure}

\subsection{Effect of Suppression Coefficient}
\label{app: beta value}

We assessed the effect of the suppression coefficient $\beta$ on the performance of the VAF method using the LLaVA-v1.5-7B model on the COCO-Random dataset within the POPE Benchmark. In our experiments, $\alpha$ was fixed at 0.15, while $\beta$ was systematically adjusted. The results, presented in \cref{fig: beta}, reveal that when $0<\beta<0.15$, VAF significantly enhanced its ability to suppress hallucinations in the model. This improvement is likely due to VAF reducing redundant attention to system prompts in this range, thereby reinforcing focus on visual features and enabling generated content to better align with the visual input. Conversely, when $\beta>0.15$, the model's performance deteriorated. We hypothesize that this decline stems from excessive suppression of attention to system prompts, which disrupts the delicate balance required for effectively integrating multimodal information, ultimately leading to a degradation in overall performance.

\begin{table*}[t]
\centering
\begin{tabular}{@{}c|c|cccc@{}}
\toprule[1pt]
\toprule
\textbf{Sampling Strategy}        & \textbf{Method}             & \textbf{Accurancy}                                           & \textbf{Precision}                                  & \textbf{Recall}                                     & \textbf{F1-Score}                                            \\ \midrule
                                  & Regular                     & 88.2                                                         & 94.4                                                & 81.4                                                & 87.4                                                         \\
\multirow{-2}{*}{Greedy}          & \cellcolor[HTML]{EFEFEF}VAF & \cellcolor[HTML]{EFEFEF}{\color[HTML]{CB0000} \textbf{89.8}} & \cellcolor[HTML]{EFEFEF}{\color[HTML]{333333} 92.9} & \cellcolor[HTML]{EFEFEF}{\color[HTML]{333333} 86.2} & \cellcolor[HTML]{EFEFEF}{\color[HTML]{CB0000} \textbf{89.4}} \\ \midrule
                                  & Regular                     & 82.9                                                         & 90.4                                                & 71.3                                                & 80.9                                                         \\
\multirow{-2}{*}{Direct Sampling} & \cellcolor[HTML]{EFEFEF}VAF & \cellcolor[HTML]{EFEFEF}{\color[HTML]{CB0000} \textbf{83.9}} & \cellcolor[HTML]{EFEFEF}{\color[HTML]{333333} 90.6} & \cellcolor[HTML]{EFEFEF}{\color[HTML]{333333} 80.9} & \cellcolor[HTML]{EFEFEF}{\color[HTML]{CB0000} \textbf{85}}   \\ \midrule
                                  & Regular                     & 84.3                                                         & 92.1                                                & 72.5                                                & 82.1                                                         \\
\multirow{-2}{*}{Top P}           & \cellcolor[HTML]{EFEFEF}VAF & \cellcolor[HTML]{EFEFEF}{\color[HTML]{CB0000} \textbf{85.7}} & \cellcolor[HTML]{EFEFEF}{\color[HTML]{333333} 89.6} & \cellcolor[HTML]{EFEFEF}{\color[HTML]{333333} 82.4} & \cellcolor[HTML]{EFEFEF}{\color[HTML]{CB0000} \textbf{85.9}} \\ \midrule
                                  & Regular                     & 83.3                                                         & 91.9                                                & 72.8                                                & 81.1                                                         \\
\multirow{-2}{*}{Top K}           & \cellcolor[HTML]{EFEFEF}VAF & \cellcolor[HTML]{EFEFEF}{\color[HTML]{CB0000} \textbf{85}}   & \cellcolor[HTML]{EFEFEF}{\color[HTML]{333333} 88.3} & \cellcolor[HTML]{EFEFEF}{\color[HTML]{333333} 81.9} & \cellcolor[HTML]{EFEFEF}{\color[HTML]{CB0000} \textbf{84.9}} \\ \midrule
                                  & Regular                     & 85.5                                                         & 95.1                                                & 74.9                                                & 84.5                                                         \\
\multirow{-2}{*}{Top K + Temp0.5} & \cellcolor[HTML]{EFEFEF}VAF & \cellcolor[HTML]{EFEFEF}{\color[HTML]{CB0000} \textbf{86.7}} & \cellcolor[HTML]{EFEFEF}{\color[HTML]{333333} 91.2} & \cellcolor[HTML]{EFEFEF}{\color[HTML]{333333} 83.4} & \cellcolor[HTML]{EFEFEF}{\color[HTML]{CB0000} \textbf{87}}   \\ \midrule
                                  & Regular                     & 80.4                                                         & 87.1                                                & 70.2                                                & 77.8                                                         \\
\multirow{-2}{*}{Top K + Temp1.5} & \cellcolor[HTML]{EFEFEF}VAF & \cellcolor[HTML]{EFEFEF}{\color[HTML]{CB0000} \textbf{82.1}} & \cellcolor[HTML]{EFEFEF}{\color[HTML]{333333} 86}   & \cellcolor[HTML]{EFEFEF}{\color[HTML]{333333} 78.2} & \cellcolor[HTML]{EFEFEF}{\color[HTML]{CB0000} \textbf{81.9}} \\ \bottomrule
\bottomrule[1pt]
\end{tabular}
\caption{\textbf{Effectiveness of the VAF method in mitigating model hallucination under different sampling strategies.} The {\color[HTML]{CB0000} \textbf{highest score}} in each setting is highlighted in {\color[HTML]{CB0000} \textbf{red}}. Experiments were conducted using the LLaVA-v1.5-7B model on the COCO-Random dataset within the POPE Benchmark.}
\label{tab: sampling VAF}
\end{table*}

\subsection{Effect of Different Sampling Strategies}
\label{app: various sampling}

We evaluated the effectiveness of the VAF method in mitigating model hallucination under different sampling strategies using the LLaVA-v1.5-7B model on the COCO-Random dataset from the POPE Benchmark. The experimental results, shown in \cref{tab: sampling VAF}, indicate that the VAF method significantly mitigates model hallucination across all sampling strategies.

\section{Prompts for Different Tasks}
\label{app: prompts}

\textbf{POPE Dataset.} In the POPE dataset, input template for the model is presented below, with the prompts highlighted in {\color[HTML]{036400}\textbf{green}} and the image highlighted in {\color[HTML]{CB0000}\textbf{red}}.
\vspace{0.2cm}
\begin{tcolorbox}[colback=gray!10, colframe=black, boxrule=0.5mm]

A chat between a curious user and an artificial intelligence assistant. The assistant gives helpful, detailed, and polite answers to the user's questions.

\vspace{0.2cm}
\begin{hangparas}{1.22cm}{1}
\textbf{USER: } {\color[HTML]{CB0000}\textbf{IMAGE}} \\
{\color[HTML]{036400}\textbf{Is there a cow in the image? Please just answer yes or no.}}
\end{hangparas}

\vspace{0.2cm}
\textbf{ASSISTANT:}
\end{tcolorbox}

\noindent \textbf{Nocaps Datasets.} In Nocaps and Flickr30k dataset, input template for the model is presented below, with prompts highlighted in {\color[HTML]{036400}\textbf{green}} and image highlighted in {\color[HTML]{CB0000}\textbf{red}}.
\vspace{0.2cm}
\begin{tcolorbox}[colback=gray!10, colframe=black, boxrule=0.5mm]

A chat between a curious user and an artificial intelligence assistant. The assistant gives helpful, detailed, and polite answers to the user's questions.

\vspace{0.2cm}
\begin{hangparas}{1.22cm}{1}
\textbf{USER: } {\color[HTML]{CB0000}\textbf{IMAGE}} \\
{\color[HTML]{036400}\textbf{Provide a one-sentence caption for the provided image.}}
\end{hangparas}

\vspace{0.2cm}
\textbf{ASSISTANT:}
\end{tcolorbox}

\vspace{0.3cm}
\noindent \textbf{Sci-VQA Dataset.} In the Sci-VQA dataset, input template for the model is presented below, with the prompts highlighted in {\color[HTML]{036400}\textbf{green}} and the image highlighted in {\color[HTML]{CB0000}\textbf{red}}. 
\vspace{0.2cm}
\begin{tcolorbox}[colback=gray!10, colframe=black, boxrule=0.5mm]

A chat between a curious user and an artificial intelligence assistant. The assistant gives helpful, detailed, and polite answers to the user's questions.

\vspace{0.2cm}
\begin{hangparas}{1.22cm}{1}
\textbf{USER: } {\color[HTML]{CB0000}\textbf{IMAGE}} \\
{\color[HTML]{036400}\textbf{Context: Select the best answer.}} \\
Which property do these three objects have in common? \\
A. shiny B. slippery C. opaque \\
{\color[HTML]{036400}\textbf{Answer with the option's letter from the given choices directly.}}
\end{hangparas}

\vspace{0.2cm}
\textbf{ASSISTANT:}

\end{tcolorbox}

%% file: main.bbl
\begin{thebibliography}{55}
\providecommand{\natexlab}[1]{#1}
\providecommand{\url}[1]{\texttt{#1}}
\expandafter\ifx\csname urlstyle\endcsname\relax
  \providecommand{\doi}[1]{doi: #1}\else
  \providecommand{\doi}{doi: \begingroup \urlstyle{rm}\Url}\fi

\bibitem[Agarwal et~al.(2020)Agarwal, Shetty, and Fritz]{agarwal2020towards}
Vedika Agarwal, Rakshith Shetty, and Mario Fritz.
\newblock Towards causal vqa: Revealing and reducing spurious correlations by invariant and covariant semantic editing.
\newblock In \emph{Proceedings of the IEEE/CVF Conference on Computer Vision and Pattern Recognition}, pages 9690--9698, 2020.

\bibitem[Agrawal et~al.(2016)Agrawal, Batra, and Parikh]{agrawal2016analyzing}
Aishwarya Agrawal, Dhruv Batra, and Devi Parikh.
\newblock Analyzing the behavior of visual question answering models.
\newblock \emph{arXiv preprint arXiv:1606.07356}, 2016.

\bibitem[Agrawal et~al.(2019)Agrawal, Desai, Wang, Chen, Jain, Johnson, Batra, Parikh, Lee, and Anderson]{Agrawal_2019}
Harsh Agrawal, Karan Desai, Yufei Wang, Xinlei Chen, Rishabh Jain, Mark Johnson, Dhruv Batra, Devi Parikh, Stefan Lee, and Peter Anderson.
\newblock nocaps: novel object captioning at scale.
\newblock In \emph{2019 IEEE/CVF International Conference on Computer Vision (ICCV)}. IEEE, 2019.

\bibitem[Bai et~al.(2023{\natexlab{a}})Bai, Bai, and et~al]{qwen}
Jinze Bai, Shuai Bai, and et al.
\newblock Qwen technical report.
\newblock \emph{arXiv preprint arXiv:2309.16609}, 2023{\natexlab{a}}.

\bibitem[Bai et~al.(2023{\natexlab{b}})Bai, Bai, and et~al]{qwenvl}
Jinze Bai, Shuai Bai, and et al.
\newblock Qwen-vl: A frontier large vision-language model with versatile abilities.
\newblock \emph{arXiv preprint arXiv:2308.12966}, 2023{\natexlab{b}}.

\bibitem[Bavishi et~al.(2023)Bavishi, Elsen, and et~al]{fuyu}
Rohan Bavishi, Erich Elsen, and et al.
\newblock Introducing our multimodal models, 2023.

\bibitem[Biten et~al.(2022)Biten, G{\'o}mez, and Karatzas]{biten2022let}
Ali~Furkan Biten, Llu{\'\i}s G{\'o}mez, and Dimosthenis Karatzas.
\newblock Let there be a clock on the beach: Reducing object hallucination in image captioning.
\newblock In \emph{Proceedings of the IEEE/CVF Winter Conference on Applications of Computer Vision}, pages 1381--1390, 2022.

\bibitem[Chen et~al.(2023{\natexlab{a}})Chen, Zhang, and et~al]{shikra}
Keqin Chen, Zhao Zhang, and et al.
\newblock Shikra: Unleashing multimodal llm's referential dialogue magic.
\newblock \emph{arXiv preprint arXiv:2306.15195}, 2023{\natexlab{a}}.

\bibitem[Chen et~al.(2023{\natexlab{b}})Chen, Sinavski, H{\"u}nermann, Karnsund, Willmott, Birch, Maund, and Shotton]{chen2023driving}
Long Chen, Oleg Sinavski, Jan H{\"u}nermann, Alice Karnsund, Andrew~James Willmott, Danny Birch, Daniel Maund, and Jamie Shotton.
\newblock Driving with llms: Fusing object-level vector modality for explainable autonomous driving.
\newblock \emph{arXiv preprint arXiv:2310.01957}, 2023{\natexlab{b}}.

\bibitem[Chen et~al.(2024)Chen, Wang, and et~al]{internvl}
Zhe Chen, Weiyun Wang, and et al.
\newblock How far are we to gpt-4v? closing the gap to commercial multimodal models with open-source suites.
\newblock \emph{arXiv preprint arXiv:2404.16821}, 2024.

\bibitem[Chiang and Li(2023)]{vicuna}
Wei-Lin Chiang and Zhuohan et~al Li.
\newblock Vicuna: An open-source chatbot impressing gpt-4 with 90\%* chatgpt quality.
\newblock \emph{See https://vicuna. lmsys. org (accessed 14 April 2023)}, 2023.

\bibitem[Dai and et~al(2023)]{instructblip}
Wenliang Dai and Junnan~Li et al.
\newblock Instructblip: Towards general-purpose vision-language models with instruction tuning, 2023.

\bibitem[Fu et~al.(2023)Fu, Chen, and et~al]{mme}
Chaoyou Fu, Peixian Chen, and et al.
\newblock Mme: A comprehensive evaluation benchmark for multimodal large language models.
\newblock \emph{arXiv preprint arXiv:2306.13394}, 2023.

\bibitem[Goyal et~al.(2017)Goyal, Khot, Summers-Stay, Batra, and Parikh]{goyal2017making}
Yash Goyal, Tejas Khot, Douglas Summers-Stay, Dhruv Batra, and Devi Parikh.
\newblock Making the v in vqa matter: Elevating the role of image understanding in visual question answering.
\newblock In \emph{Proceedings of the IEEE conference on computer vision and pattern recognition}, pages 6904--6913, 2017.

\bibitem[Gunjal et~al.(2023)Gunjal, Yin, and Bas]{gunjal2023detecting}
Anisha Gunjal, Jihan Yin, and Erhan Bas.
\newblock Detecting and preventing hallucinations in large vision language models.
\newblock \emph{arXiv preprint arXiv:2308.06394}, 2023.

\bibitem[Gupta et~al.(2022)Gupta, Li, Kortylewski, Zhang, Li, and Yuille]{gupta2022swapmix}
Vipul Gupta, Zhuowan Li, Adam Kortylewski, Chenyu Zhang, Yingwei Li, and Alan Yuille.
\newblock Swapmix: Diagnosing and regularizing the over-reliance on visual context in visual question answering.
\newblock In \emph{Proceedings of the IEEE/CVF Conference on Computer Vision and Pattern Recognition}, pages 5078--5088, 2022.

\bibitem[Han et~al.(2022)Han, Nie, Yin, Wu, and Yan]{han2022visual}
Yudong Han, Liqiang Nie, Jianhua Yin, Jianlong Wu, and Yan Yan.
\newblock Visual perturbation-aware collaborative learning for overcoming the language prior problem.
\newblock \emph{arXiv preprint arXiv:2207.11850}, 2022.

\bibitem[Hu et~al.(2023)Hu, Pan, Li, and Yang]{hu2023advancing}
Mingzhe Hu, Shaoyan Pan, Yuheng Li, and Xiaofeng Yang.
\newblock Advancing medical imaging with language models: A journey from n-grams to chatgpt.
\newblock \emph{arXiv preprint arXiv:2304.04920}, 2023.

\bibitem[Huang et~al.(2024)Huang, Dong, Zhang, Wang, He, Wang, Lin, Zhang, and Yu]{opera}
Qidong Huang, Xiaoyi Dong, Pan Zhang, Bin Wang, Conghui He, Jiaqi Wang, Dahua Lin, Weiming Zhang, and Nenghai Yu.
\newblock Opera: Alleviating hallucination in multi-modal large language models via over-trust penalty and retrospection-allocation.
\newblock In \emph{CVPR}, pages 13418--13427, 2024.

\bibitem[Hudson and Manning(2019)]{gqa}
Drew~A Hudson and Christopher~D Manning.
\newblock Gqa: A new dataset for real-world visual reasoning and compositional question answering.
\newblock In \emph{CVPR}, pages 6700--6709, 2019.

\bibitem[Huo et~al.(2024)Huo, Xu, Zhang, Wang, Chen, and Zhao]{huo2024selfintrospectivedecodingalleviatinghallucinations}
Fushuo Huo, Wenchao Xu, Zhong Zhang, Haozhao Wang, Zhicheng Chen, and Peilin Zhao.
\newblock Self-introspective decoding: Alleviating hallucinations for large vision-language models, 2024.

\bibitem[Jiang et~al.(2024)Jiang, Xu, and et~al]{hacl}
Chaoya Jiang, Haiyang Xu, and et al.
\newblock Hallucination augmented contrastive learning for multimodal large language model.
\newblock In \emph{CVPR}, pages 27036--27046, 2024.

\bibitem[Leng et~al.(2024)Leng, Zhang, and et~al]{vcd}
Sicong Leng, Hang Zhang, and et al.
\newblock Mitigating object hallucinations in large vision-language models through visual contrastive decoding.
\newblock In \emph{CVPR}, pages 13872--13882, 2024.

\bibitem[Li et~al.(2023{\natexlab{a}})Li, Zhang, and et~al]{mimic}
Bo Li, Yuanhan Zhang, and et al.
\newblock Mimic-it: Multi-modal in-context instruction tuning.
\newblock \emph{arXiv preprint arXiv:2306.05425}, 2023{\natexlab{a}}.

\bibitem[Li et~al.(2024)Li, Zhang, and et~al]{llavanext}
Bo Li, Kaichen Zhang, and et al.
\newblock Llava-next: Stronger llms supercharge multimodal capabilities in the wild, 2024.

\bibitem[Li et~al.(2023{\natexlab{b}})Li, Wong, and et~al]{llavamed}
Chunyuan Li, Cliff Wong, and et al.
\newblock Llava-med: Training a large language-and-vision assistant for biomedicine in one day.
\newblock In \emph{NeurIPS}, pages 28541--28564, 2023{\natexlab{b}}.

\bibitem[Li et~al.(2023{\natexlab{c}})Li, Li, Savarese, and Hoi]{blip2}
Junnan Li, Dongxu Li, Silvio Savarese, and Steven Hoi.
\newblock {BLIP}-2: Bootstrapping language-image pre-training with frozen image encoders and large language models.
\newblock In \emph{ICML}, 2023{\natexlab{c}}.

\bibitem[Li et~al.(2022)Li, Holtzman, and et~al]{contrastive}
Xiang~Lisa Li, Ari Holtzman, and et al.
\newblock Contrastive decoding: Open-ended text generation as optimization.
\newblock \emph{arXiv preprint arXiv:2210.15097}, 2022.

\bibitem[Li et~al.(2023{\natexlab{d}})Li, Du, Zhou, Wang, Zhao, and Wen]{li2023evaluating}
Yifan Li, Yifan Du, Kun Zhou, Jinpeng Wang, Wayne~Xin Zhao, and Ji-Rong Wen.
\newblock Evaluating object hallucination in large vision-language models.
\newblock \emph{arXiv preprint arXiv:2305.10355}, 2023{\natexlab{d}}.

\bibitem[Li et~al.(2023{\natexlab{e}})Li, Du, Zhou, Wang, Zhao, and Wen]{pope}
Yifan Li, Yifan Du, Kun Zhou, Jinpeng Wang, Xin Zhao, and Ji-Rong Wen.
\newblock Evaluating object hallucination in large vision-language models.
\newblock In \emph{EMNLP}, pages 292--305, 2023{\natexlab{e}}.

\bibitem[Lin et~al.(2014)Lin, Maire, and et~al]{mscoco}
Tsung-Yi Lin, Michael Maire, and et al.
\newblock Microsoft coco: Common objects in context.
\newblock In \emph{ECCV}, pages 740--755, 2014.

\bibitem[Liu et~al.(2023{\natexlab{a}})Liu, Lin, Li, Wang, Yacoob, and Wang]{liu2023mitigating}
Fuxiao Liu, Kevin Lin, Linjie Li, Jianfeng Wang, Yaser Yacoob, and Lijuan Wang.
\newblock Mitigating hallucination in large multi-modal models via robust instruction tuning.
\newblock \emph{arXiv preprint arXiv:2306.14565}, 2023{\natexlab{a}}.

\bibitem[Liu et~al.(2023{\natexlab{b}})Liu, Li, Wu, and Lee]{llava}
Haotian Liu, Chunyuan Li, Qingyang Wu, and Yong~Jae Lee.
\newblock Visual instruction tuning.
\newblock In \emph{NeurIPS}, pages 34892--34916, 2023{\natexlab{b}}.

\bibitem[Liu et~al.(2023{\natexlab{c}})Liu, Zhu, Kato, Kondo, Aoyama, and Hasegawa]{liu2023llm}
Haokun Liu, Yaonan Zhu, Kenji Kato, Izumi Kondo, Tadayoshi Aoyama, and Yasuhisa Hasegawa.
\newblock Llm-based human-robot collaboration framework for manipulation tasks.
\newblock \emph{arXiv preprint arXiv:2308.14972}, 2023{\natexlab{c}}.

\bibitem[Liu et~al.(2024{\natexlab{a}})Liu, Li, Li, and Lee]{llava1.5}
Haotian Liu, Chunyuan Li, Yuheng Li, and Yong~Jae Lee.
\newblock Improved baselines with visual instruction tuning.
\newblock In \emph{CVPR}, pages 26296--26306, 2024{\natexlab{a}}.

\bibitem[Liu et~al.(2024{\natexlab{b}})Liu, Courant, and Kalogeiton]{funny}
Zhi-Song Liu, Robin Courant, and Vicky Kalogeiton.
\newblock Funnynet-w: Multimodal learning of funny moments in videos in the wild.
\newblock \emph{International Journal of Computer Vision}, pages 1--22, 2024{\natexlab{b}}.

\bibitem[Lovenia et~al.(2023)Lovenia, Dai, Cahyawijaya, Ji, and Fung]{lovenia2023negative}
Holy Lovenia, Wenliang Dai, Samuel Cahyawijaya, Ziwei Ji, and Pascale Fung.
\newblock Negative object presence evaluation (nope) to measure object hallucination in vision-language models.
\newblock \emph{arXiv preprint arXiv:2310.05338}, 2023.

\bibitem[Lu et~al.(2022)Lu, Mishra, Xia, Qiu, Chang, Zhu, Tafjord, Clark, and Kalyan]{lu2022learn}
Pan Lu, Swaroop Mishra, Tony Xia, Liang Qiu, Kai-Wei Chang, Song-Chun Zhu, Oyvind Tafjord, Peter Clark, and Ashwin Kalyan.
\newblock Learn to explain: Multimodal reasoning via thought chains for science question answering.
\newblock In \emph{The 36th Conference on Neural Information Processing Systems (NeurIPS)}, 2022.

\bibitem[Mai et~al.(2023)Mai, Chen, Li, Qian, Elhoseiny, and Ghanem]{mai2023llm}
Jinjie Mai, Jun Chen, Bing Li, Guocheng Qian, Mohamed Elhoseiny, and Bernard Ghanem.
\newblock Llm as a robotic brain: Unifying egocentric memory and control.
\newblock \emph{arXiv preprint arXiv:2304.09349}, 2023.

\bibitem[Meta(2024)]{llama3}
AI Meta.
\newblock Introducing meta llama 3: The most capable openly available llm to date.
\newblock \emph{Meta AI}, 2024.

\bibitem[Niu et~al.(2021)Niu, Tang, Zhang, Lu, Hua, and Wen]{niu2021counterfactual}
Yulei Niu, Kaihua Tang, Hanwang Zhang, Zhiwu Lu, Xian-Sheng Hua, and Ji-Rong Wen.
\newblock Counterfactual vqa: A cause-effect look at language bias.
\newblock In \emph{Proceedings of the IEEE/CVF Conference on Computer Vision and Pattern Recognition}, pages 12700--12710, 2021.

\bibitem[Schwenk et~al.(2022)Schwenk, Khandelwal, and et~al]{aokvqa}
Dustin Schwenk, Apoorv Khandelwal, and et al.
\newblock A-okvqa: A benchmark for visual question answering using world knowledge.
\newblock In \emph{ECCV}, pages 146--162, 2022.

\bibitem[Taori et~al.(2023)Taori, Gulrajani, and et~al]{stanford}
Rohan Taori, Ishaan Gulrajani, and et al.
\newblock Stanford alpaca: an instruction-following llama model (2023).
\newblock \emph{URL https://github. com/tatsu-lab/stanford\_alpaca}, 1\penalty0 (9), 2023.

\bibitem[Touvron et~al.(2023{\natexlab{a}})Touvron, Lavril, and et~al]{llama}
Hugo Touvron, Thibaut Lavril, and et al.
\newblock Llama: Open and efficient foundation language models.
\newblock \emph{arXiv preprint arXiv:2302.13971}, 2023{\natexlab{a}}.

\bibitem[Touvron et~al.(2023{\natexlab{b}})Touvron, Martin, and et~al]{llama2}
Hugo Touvron, Louis Martin, and et al.
\newblock Llama 2: Open foundation and fine-tuned chat models.
\newblock \emph{arXiv preprint arXiv:2307.09288}, 2023{\natexlab{b}}.

\bibitem[Wang et~al.(2023)Wang, Zhao, Ouyang, Wang, and Shen]{wang2023chatcad}
Sheng Wang, Zihao Zhao, Xi Ouyang, Qian Wang, and Dinggang Shen.
\newblock Chatcad: Interactive computer-aided diagnosis on medical image using large language models.
\newblock \emph{arXiv preprint arXiv:2302.07257}, 2023.

\bibitem[Wang et~al.(2024)Wang, Pan, and et~al]{icd}
Xintong Wang, Jingheng Pan, and et al.
\newblock Mitigating hallucinations in large vision-language models with instruction contrastive decoding.
\newblock \emph{arXiv preprint arXiv:2403.18715}, 2024.

\bibitem[Wu et~al.(2022)Wu, Zhao, Zhao, Zhang, Yuan, Zhao, and Jiang]{wu2022overcoming}
Yike Wu, Yu Zhao, Shiwan Zhao, Ying Zhang, Xiaojie Yuan, Guoqing Zhao, and Ning Jiang.
\newblock Overcoming language priors in visual question answering via distinguishing superficially similar instances.
\newblock \emph{arXiv preprint arXiv:2209.08529}, 2022.

\bibitem[Wu et~al.(2023)Wu, Wang, Xu, Lu, and Yan]{wu2023embodied}
Zhenyu Wu, Ziwei Wang, Xiuwei Xu, Jiwen Lu, and Haibin Yan.
\newblock Embodied task planning with large language models.
\newblock \emph{arXiv preprint arXiv:2307.01848}, 2023.

\bibitem[Yan et~al.(2023)Yan, Liu, Feng, and Huang]{yan2023overcoming}
Hong Yan, Lijun Liu, Xupeng Feng, and Qingsong Huang.
\newblock Overcoming language priors with self-contrastive learning for visual question answering.
\newblock \emph{Multimedia Tools and Applications}, 82\penalty0 (11):\penalty0 16343--16358, 2023.

\bibitem[Ye et~al.(2023)Ye, Xu, and et~al]{mplug}
Qinghao Ye, Haiyang Xu, and et al.
\newblock mplug-owl: Modularization empowers large language models with multimodality.
\newblock \emph{arXiv preprint arXiv:2304.14178}, 2023.

\bibitem[Zhang et~al.(2023)Zhang, Sun, and et~al]{gpt4roi}
Shilong Zhang, Peize Sun, and et al.
\newblock Gpt4roi: Instruction tuning large language model on region-of-interest.
\newblock \emph{arXiv preprint arXiv:2307.03601}, 2023.

\bibitem[Zhibo et~al.(2023)Zhibo, Huizhen, Muhua, Yichao, Tong, and Jingbo]{zhibo2023overcoming}
Ren Zhibo, Wang Huizhen, Zhu Muhua, Wang Yichao, Xiao Tong, and Zhu Jingbo.
\newblock Overcoming language priors with counterfactual inference for visual question answering.
\newblock In \emph{Proceedings of the 22nd Chinese National Conference on Computational Linguistics}, pages 600--610, 2023.

\bibitem[Zhou et~al.(2022)Zhou, Yang, Loy, and Liu]{coop}
Kaiyang Zhou, Jingkang Yang, Chen~Change Loy, and Ziwei Liu.
\newblock Learning to prompt for vision-language models.
\newblock \emph{IJCV}, 130\penalty0 (9):\penalty0 2337--2348, 2022.

\bibitem[Zhu et~al.(2023)Zhu, Chen, Shen, Li, and Elhoseiny]{minigpt}
Deyao Zhu, Jun Chen, Xiaoqian Shen, Xiang Li, and Mohamed Elhoseiny.
\newblock Minigpt-4: Enhancing vision-language understanding with advanced large language models.
\newblock \emph{arXiv preprint arXiv:2304.10592}, 2023.

\end{thebibliography}
